**Enhancing EBSD throughput of battery electrode materials using super-resolution generative adversarial networks**

John Mangum[1], Andrew Glaws[1], Francois Usseglio-Viretta[1], Steven Spurgeon[1], Donal P. Finegan[1]

[1] National Laboratory of the Rockies (NLR), 15013 Denver W Parkway, Golden, CO 80401, USA

# Abstract

Quantitative microstructural characterization of Li-ion battery electrode materials using electron backscatter diffraction (EBSD) has been proven as a critical method for optimizing cell performance. However, the inherently slow nature of EBSD can hinder the throughput of analyses needed for statistical representation of a material microstructure being developed. This work demonstrates a machine learning super-resolution framework using a generative adversarial network (SRGAN) to significantly increase EBSD throughput. The SRGAN model was trained on EBSD data of $LiNi_xMn_yCo_zO_2$ (NMC) cathode particles to computationally enhance low-resolution datasets and its performance is compared against classical interpolation methods across various upscaling factors (2× to 12×).

Both qualitative image metrics and quantitative microstructural analysis verified that the SRGAN systematically outperformed classical methods, particularly in preserving small grains and maintaining realistic grain boundaries. We demonstrate that a 5× upscaling factor, corresponding to a 25× speed-up in acquisition time or a 25× larger field of view, is practical while maintaining acceptable accuracy in key metrics like grain size and shape. For instance, at 5× upscaling, relative errors were +5.7%, +8.2%, and -14.6% on grain area-equivalent diameter, grain maximum sphere-inscribed diameter, and grain boundary length, respectively. The SRGAN methodology developed in this work significantly enhances the efficiency of EBSD acquisition for more statistically robust microstructural dataset, enabling EBSD as a high-throughput characterization tool for materials research and industrial process development.

# Introduction

Lithium (Li) -ion cells are pivotal for energy-storage systems spanning portable electronics and grid-scale storage. Since Li-ion cells were first commercialized in the early 1990's, there has been tremendous progress in increasing their energy-density, power-density, and cycle-life, much of which was achieved by incremental improvements to the chemical, crystallographic, and morphological properties of active materials. Most positive electrodes that are used in the field today for applications like electric vehicles consist of polycrystalline particles with layered crystal structures such as $LiNi_xMn_yCo_zO_2$ (NMC). The morphology and orientations of the crystals within those particles have been shown to greatly influence their performance. This is because each grain facilitates directional Li-ion transport along 2-dimensional planes. Relative orientation between grains determines the tortuosity of the sub-particle Li transport during intercalation and deintercalation, and the crystal's anisotropic volume changes during operation also create complex strain profiles that can lead to particle cracking, shortening the cycle life of the cell.[1–3] Thus, control of the grain orientation presents an opportunity for improving the rate capability and cycle life of cells. For example, Ren et al.[4] showed dramatic increase in rate capability of NMC particles by radially aligning particles (e.g. from ~30 mAh/g to ~135 mAh/g at a discharge rate of 20C) and Kim et al.[5] showed that by radially aligning elongated grains and by selectively doping the crystal structure with other transition metals, the sub-particle strain profiles and consequent particle cracking can be suppressed, extending the life of the cathode considerably. Crack mitigation can also be achieved by manufacturing single crystal particles,[6] but the rate capability of single crystal particles has been shown to be inferior to their polycrystalline counterparts.[7]

The clear influence of sub-particle grain morphology and orientation of cell performance motivates further research and manufacturing optimization of grain properties. However, to achieve quantitative relationships between grain properties and cell performance, it is essential to be able to quantitatively measure grain shapes and orientation, not just at the single particle level but across many particles from a synthesized batch to achieve statistical representation. Electron backscatter diffraction (EBSD) has emerged as the most promising method to measure grain properties and support development of positive electrode particles; this is due to it being a highly accessible lab-based technique that supports collection of high-resolution spatially-resolved crystallographic properties, including morphology, crystal orientation, as well as other metrics like misorientation and strain for many particles.[8–10] The use of focused-ion beam (FIB)-EBSD also enables extending the spatial mapping of sub-particle crystal properties into 3D.[11] However, being a rastering technique, where each pixel in a 2D inverse pole figure (IPF) requires a distinct EBSD measurement, the technique is relatively slow and collecting data across many particles for statistical representativeness is hindered. This poses a challenge for research and development, where high throughput automated measurements are needed to keep pace with material development effort. Furthermore, FIB-EBSD to acquire 3D data that is needed for important transport and morphological metrics in particles is even more time-constrained than simple 2D measurements, particularly considering that achieving statistical representativeness may require hundreds of particles. Even assuming the rollout of 2D to 3D generative modeling[12,13] and large language models,[14] a substantial amount of 2D IPF data is still expected to be needed. However, 2D to 3D generation is expected to be key to enabling statistical representation in particle metrics which will require acquisition of 2D data within a favorable time-frame (e.g. less than half a day to characterize a batch of particles). Accelerating the acquisition of EBSD across 2D cross sections of many particles is the first step towards achieving this. High-throughput EBSD would facilitate better statistics, the possibility of generating representative 3D particles, and would enable faster feedback during material development stages that could accelerate advanced manufacturing. Faster image acquisition than what today's hardware alone can achieve is needed, which necessitates solutions in software.

Super-resolution refers to the process of computationally enhancing image data to increase the pixel resolution, ideally by injecting meaningful sub-pixel details to the image.[15] Classical approaches to super-resolution (e.g., nearest neighbor, linear/cubic interpolation, etc.) are computationally efficient and do not require any training data. However, these highly general approaches typically result in overly smoothed images that struggle to capture critical small-scale features in the enhanced image. More recently, machine learning methods have provided powerful new capabilities for performing image super-resolution. Specifically, deep learning models comprised of convolutional neural networks (CNNs) constitute the state of the art for image super-resolution due to their ability to effectively process image-based data and learn complex mappings between low-resolution (LR) inputs and their high-resolution (HR) counterparts. While simple CNN architectures were originally used for these models,[16,17] research has also explored the use of more sophisticated frameworks, such as generative adversarial networks (GANs)[18,19] and diffusion models.[20]

The original application of these super-resolution models focused on photographic images from large-scale curated datasets (e.g., ImageNet[21]) that generally include images of objects, animals, cars, etc. Follow on work has explored the use of super-resolution for scientific data, such as climate and weather data,[22,23] medical imaging,[24,25] astronomical images,[26,27] and microscopy. Research into data-driven super-resolving microscopy data has focused primarily on the SEM imaging.[28,29] For EBSD imaging, physics-based approaches have been proposed to enhance orientation maps.[30] Jung, et al. explored the use of a super-resolution residual network (SRResNet) model to enhance grain orientation and phase data from EBSD imaging.[31] They train models to perform 4x, 8x, and 16x super-resolution on synthetic datasets and examine the results of finite element simulations using true high-resolution and super-resolved data. However, their work does not consider more powerful deep learning approaches, such as GANs, which have been demonstrated to outperform models trained using traditional losses. Additionally, these models were not trained on real datasets that include challenges that would arise in real imaging process, such as noise and stage drift.

In this work, we explore the use of GANs for super-resolution to accelerate EBSD analysis by enabling the use of coarser imaging processes with minimal impact to the overall quality of the results. Specifically, this work examines questions related to the trade-offs between increasing the degree of synthetic resolution enhancement and the quality of the resulting image and microstructural analysis. To do this, we train super-resolution GAN (SRGAN) models to perform from 2× to 12× pixel resolution enhancement and study the quality of the resulting images. Note that a 4× coarsening of the imaging process results in a 16× speed-up in the overall imaging time. The super-resolution is performed on a combination of band contrast and labeled boundary data, requiring a mixed modeling approach to handle both grayscale and binary outputs simultaneously. We validate the performance of the trained SRGANs for various resolution enhancement across both pixel-based and grain-based metrics and demonstrate a 25× increase in acquisition rate is possible without losing important detail in quantitative microstructural metrics.

## Study overview and general methodology

The goal of this work is to examine the performance and trade-offs of SRGAN-based data enhancement of EBSD images of NMC particles across a range of resolution enhancement levels. Figure 1 provides a general overview of the data collected and studies performed in this work. EBSD data was gathered for 15 cross-sectioned particles at two different resolutions. The high-resolution imaging used 25 nm step sizes (pixel size), and the low-resolution was imaged using 100 nm step sizes. The high-resolution imaging took ~16× the time as the low-resolution imaging (i.e., 16 hours vs. 1 hour). To obtain training and testing data across multiple resolution enhancements, we artificially downscaled, or coarsened, the high-resolution data to varying degrees. The imaged 1/4 resolution data was used as a point of comparison for this downscaling procedure. The low-resolution data was then upscaled, or enhanced, using the SRGAN approach as well as multiple classical approaches. The quality of the enhanced data was examined by comparing to the imaged high-resolution data. The remainder of this section discusses the background and technical details for each of these steps before we explore the results in the next section.

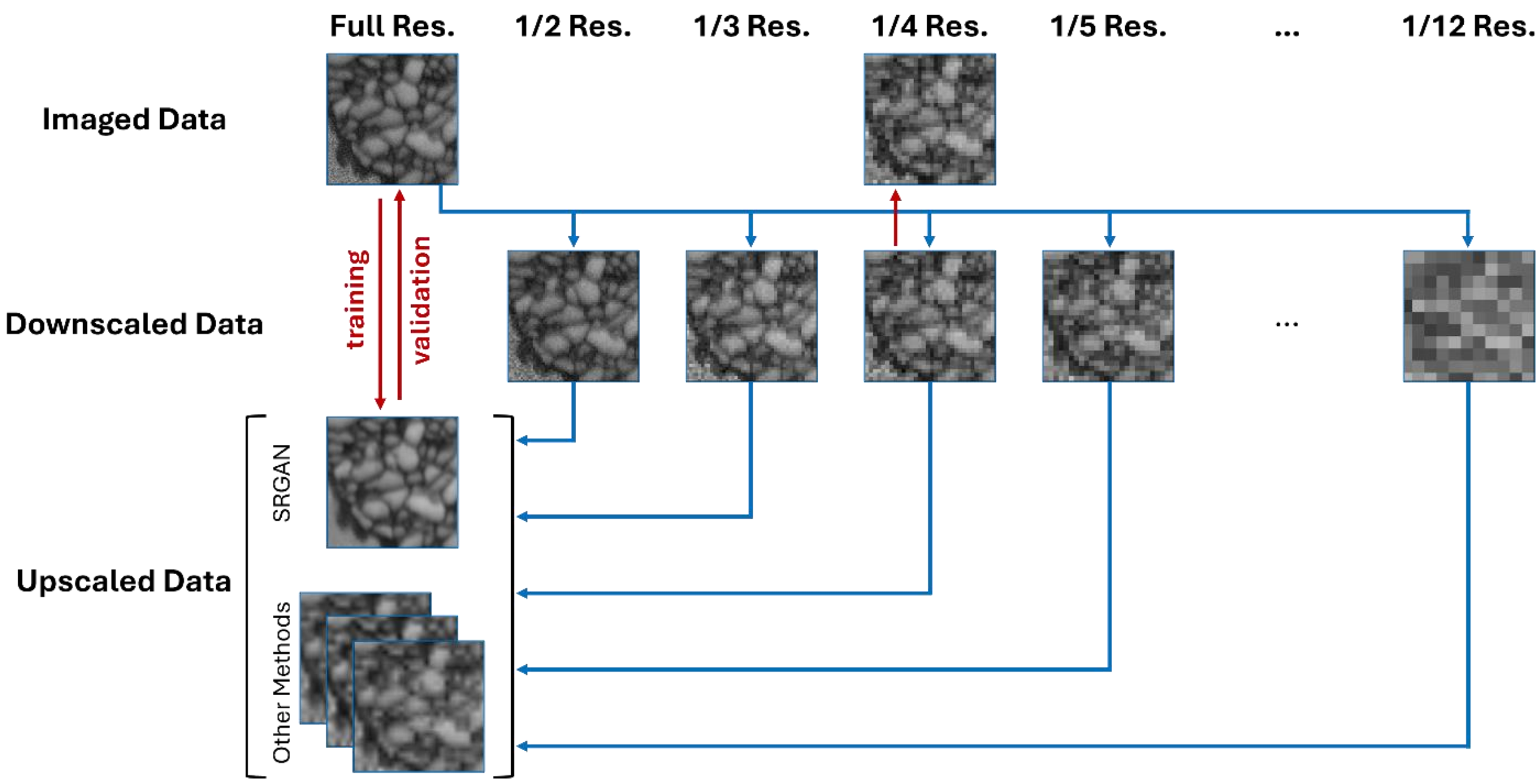

Figure 1. A general schematic of the data and processes used in this study. Ground truth imaged data were acquired at full and ¼ resolution. Downscaled data were generated from the full resolution with downscaling factors ranging from 2× to 12×. All downscaled data, as well as the imaged ¼ resolution data, were subsequently upscaled back to full resolution using various methods.

# Experimental data

## Electrode materials:

The electrode used in this work consisted of Targray $LiNi_{0.8}Mn_{0.1}Co_{0.1}O_2$ (NMC811) particles. The NMC811 was mixed with Timcal C45 carbon, and solvay 5130 PVDF binder to the weight percentages of 90 wt%, 5 wt%, and 5 wt% respectively. The electrodes were coated to achieve 9.08 mg $cm^{-2}$ and calendered to a coating thickness of ~53 μm with a porosity of ~33%. The electrodes were prepared by the Cell Analysis, Modeling, and Prototyping (CAMP) Facility at Argonne National Laboratory.

## Sample preparation and imaging:

NMC811 electrode samples were cut down to approximately 1 $cm^2$ using scissors and then adhered to a similarly sized Si wafer, to act as a rigid handle, using conductive Ag paint (Ted Pella Inc, Prod. #16062). Smooth cross-sectional faces were then obtained by polishing one edge of the electrode using a JEOL broad beam Ar cross-section polisher (IB-19550CCP) for up to 6 hours with 5 kV Ar ions and ~120 μA current at ambient temperature.

Electron backscatter diffraction (EBSD) of the polished cross-sectional faces was acquired using an Oxford Instruments Symmetry EBSD detector mounted on a FEI Nova NanoSEM 630. The samples were mounted on a 20° pre-tilt holder such that the cross-sectional face was pre-tilted to 70° and aligned facing the EBSD detector. An accelerating voltage of 20 kV and beam current of 6.4 nA was used with the sample set at approximately 11 mm working distance. Oxford Instruments AZtec software was used to acquire the EBSD data and export the processed map data (band contrast, Euler angles) and metadata in an HDF5 file format (.h5oina) for further processing and use in training and validating the super-resolution framework. Oxford Instruments AZtecCrystal software was also used to clean up the map data using standard methods and export grain boundary maps.

# Numerical methods

## Image resolution mapping

### *Super resolution generative adversarial network*

In this work, we implement a super-resolution generative adversarial network (SRGAN) framework to enhance low-resolution EBSD data. In general, the GAN approach trains two competing networks:  a generator $G$ that creates new data and a discriminator $D$ that attempts to distinguish between real and synthetic data samples.[32] When performing super-resolution, the generator takes in low-resolution input data and maps it to a corresponding high-resolution output image, $G: \mathcal{X}_{low} \to \mathcal{X}_{high}$, where $\mathcal{X}_{low}$ and $\mathcal{X}_{high}$ represent the space of low- and high-resolution images, respectively. The generator model is trained on a combination of two loss functions: the content loss and the adversarial loss. The content loss compares the generated and real high-resolution images in a pixel-wise manner. Here, we are performing super-resolution on images containing both band contrast and boundary data, which mixes continuous and binary data types. To accommodate this, our content loss combines the mean squared error over the band contrast data with the binary cross entropy loss over the boundary data. The adversarial loss is computed by feedback from the discriminator model. The discriminator is provided with examples of real high-resolution images and artificial images generated by $G$ and attempts to classify them appropriately. The adversarial training process drives $D$ to better distinguish between real and fake data and $G$ to produce more realistic images. Mathematically, this is expressed as a minmax optimization,

$$\min_G \max_D \left[\mathbb{E} \log D(x_{HR}) + \mathbb{E} \log\left(1 - D\big(G(x_{LR})\big)\right)\right] \quad [1]$$

where $x_{LR}$ and $x_{HR}$ are low- and high-resolution images. Figure 2 provides a schematic overview of this training process.

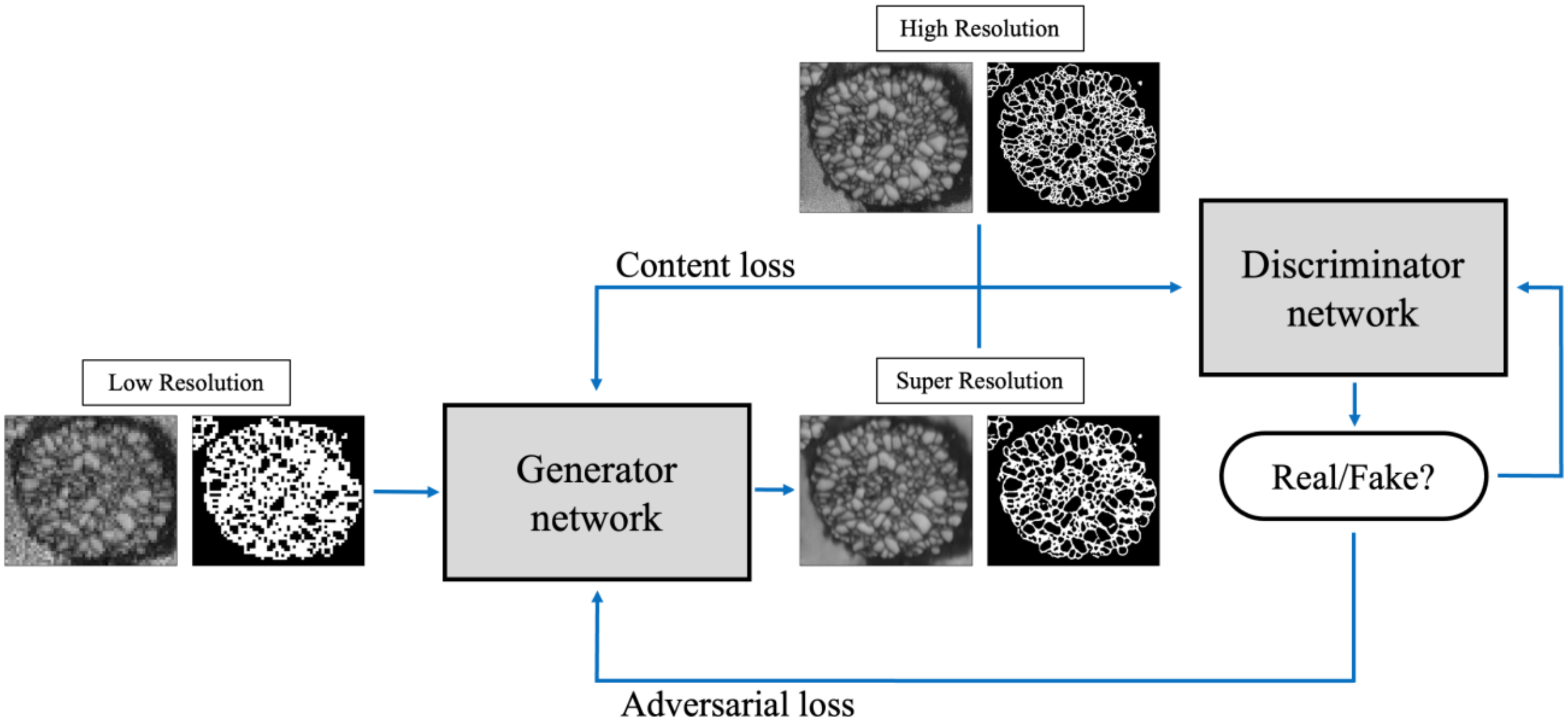

Figure 2. Process flow showing how low-resolution data is super-resolved and compared against the ground truth high-resolution data to train the SRGAN model.

For this work, the generator model contains 10 convolutional layers with a 3x3 kernel, step size of one, and hyperbolic tangent activations. The resolution is enhanced the appropriate amount using a depth-to-width pixel shuffle layer. The discriminator model contains 7 convolutional layers with the same properties as the generator. The processed image is then flattened and fed through three fully-connected layers with widths of 128, 64, and 1 to predict whether or not a given image is real or generated. The model parameters are updated in an alternating fashion using the Adam optimizer.[33]

## *Image downscaling:*

One critical feature of SRGAN models is their reliance on representative training data. That is, the model requires corresponding pairs of low-resolution and high-resolution data to learn a statistical mapping between them. In this work, we seek to explore the trade-offs between the degree of data enhancement (upscaling factor) and the quality of the generated data by training SRGAN models to perform resolution enhancements ranging from 2× to 12× upscaling factors. To avoid collecting paired low- and high-resolution images across all levels of enhancement studied here, we look to artificially downscale, or coarsen, the ground truth high-resolution image data. A common approach in SRGANs literature is to convolutionally average blocks of high-resolution pixels to obtain low-resolution counterparts. However, in the context of EBSD imaging, this procedure does not accurately reflect the nature of collecting coarse data. Instead, reducing acquisition resolution (i.e., increasing step size) merely increases the spacing between acquired pixels without incorporating averaged information from surrounding pixels. To more realistically mimic this process, we take a centered 2x2 or 3x3 block (for even or odd resolution enhancements, respectively) and randomly select a pixel from this block. Figure 3a provides an overview of this process. In Figure 3b, we examine the quality of the downscaling process the imaged low-resolution data. The figure contains a semivariogram, which measures the relative variance in an image as a function of the distance from each pixel. In this plot, we can see that the block-averaged data overly effectively smooths the downscaled imaged, which would result in the SRGAN model learning an incorrect relationship between low- and high-resolution data. The randomly sampled approach matches the level of variance we see in the imaged low-resolution data.

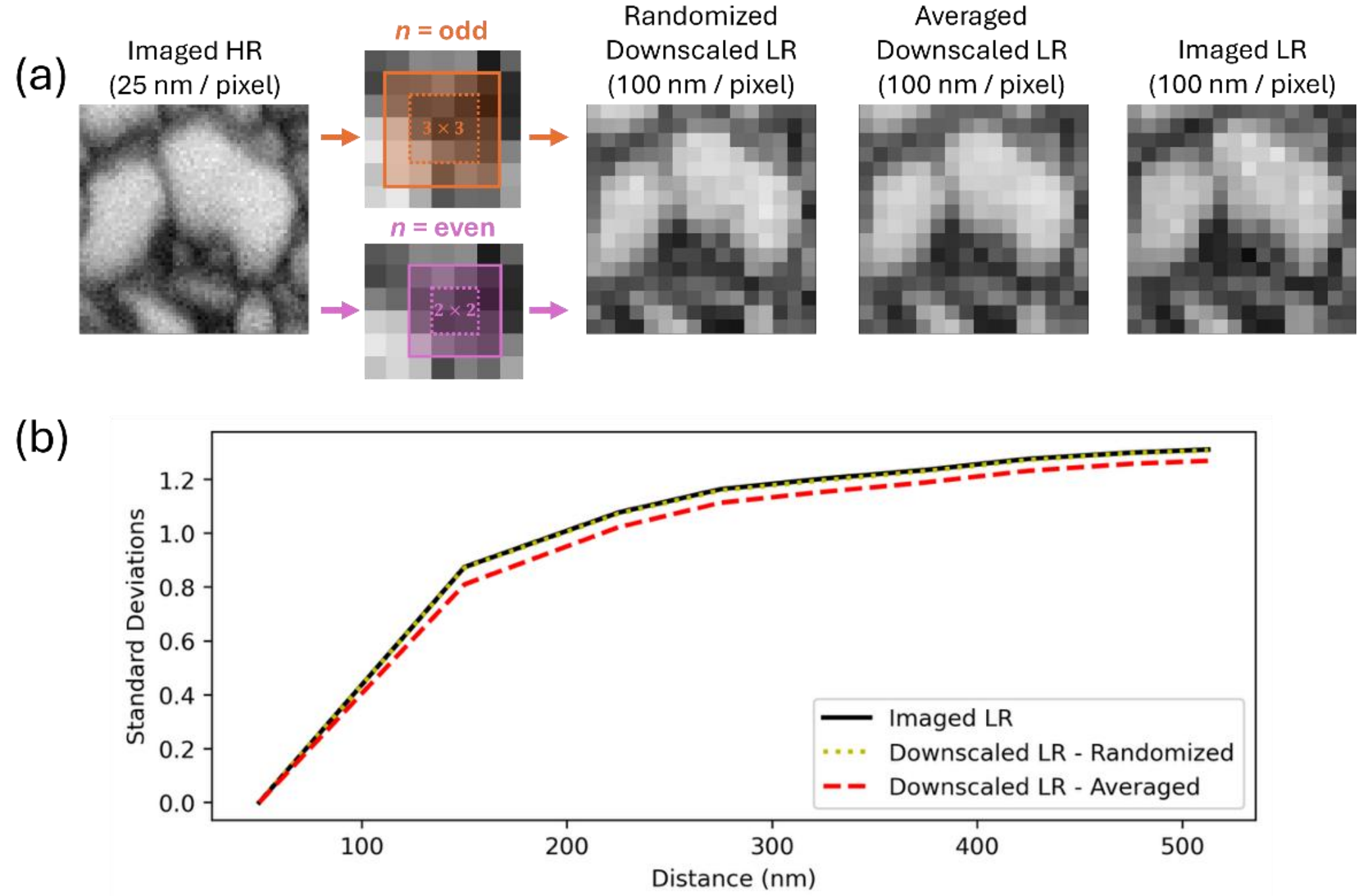

Figure 3. (a) Illustration qualitatively showing the artificial downscaling methods and comparison to imaged true low-resolution data. (b) Semivariogram showing the average variance of each low-resolution image as a function of distance from each pixel. Averaged downscaling artificially smooths the resulting low-resolution data and does not as accurately represent the true low-resolution data.

## Grains and grain boundaries identification:

Grains and grain boundaries have been segmented starting from the raw grain boundary map (cf. Fig. 4a). The workflow is illustrated in Figure 4 and explained here.

The field of view (FOV) may contain several particles. To avoid truncated particles that bias the particle size and shape analysis, only the particle located at the FOV's center must be kept. Therefore, a particle identification workflow is applied. First, the background domain is identified. To do this, the clusters of the complementary domain of the grain boundaries are identified (cf. Fig. 4b) using the MATLAB built-in function *bwlabel*. Then, the truncated clusters are removed. The union of the grain boundaries and of these removed clusters represent the background (in grey, cf. Fig. 4c). Second, the main particle centered in the FOV is identified. The particle domain is retrieved dilating the grains using the sphere-based local dilation method (cf., Fig. 4d).[34] This method ensures the exterior boundary of the particles present in the FOV are not dilated (unlike morphology closing[35]), which prevents creating irrelevant connections between nearby particles. Then, the largest cluster is identified as the particle of interest (cf. Fig. 4e). Particle shape (circularity and solidity) and size (area-equivalent diameter) are then calculated. The grains of the particle of interest (cf. Fig. 4f) are then identified multiplying the grain cluster map (cf. Fig. 4b) with the particle binary image (i.e., Fig. 4e). The grain boundaries of the particle of interest (cf. Fig. 4g) are then identified multiplying the grain boundary map (cf. Fig. 4a) with the particle binary image (i.e., Fig. 4e). However, the raw boundary map shows variable widths with some nearby grain boundaries being merged together due to the limited image resolution. This prevents calculating the length of the grain boundaries. To remedy this, the grain boundary image is skeletonize using the MATLAB built-in function *bwskel*, which is based on the medial axis transform.[36,37] Coarse boundaries are either converted to several interfaces or to a single intersection (respectively, red and green rectangles in Figures 4g and h).

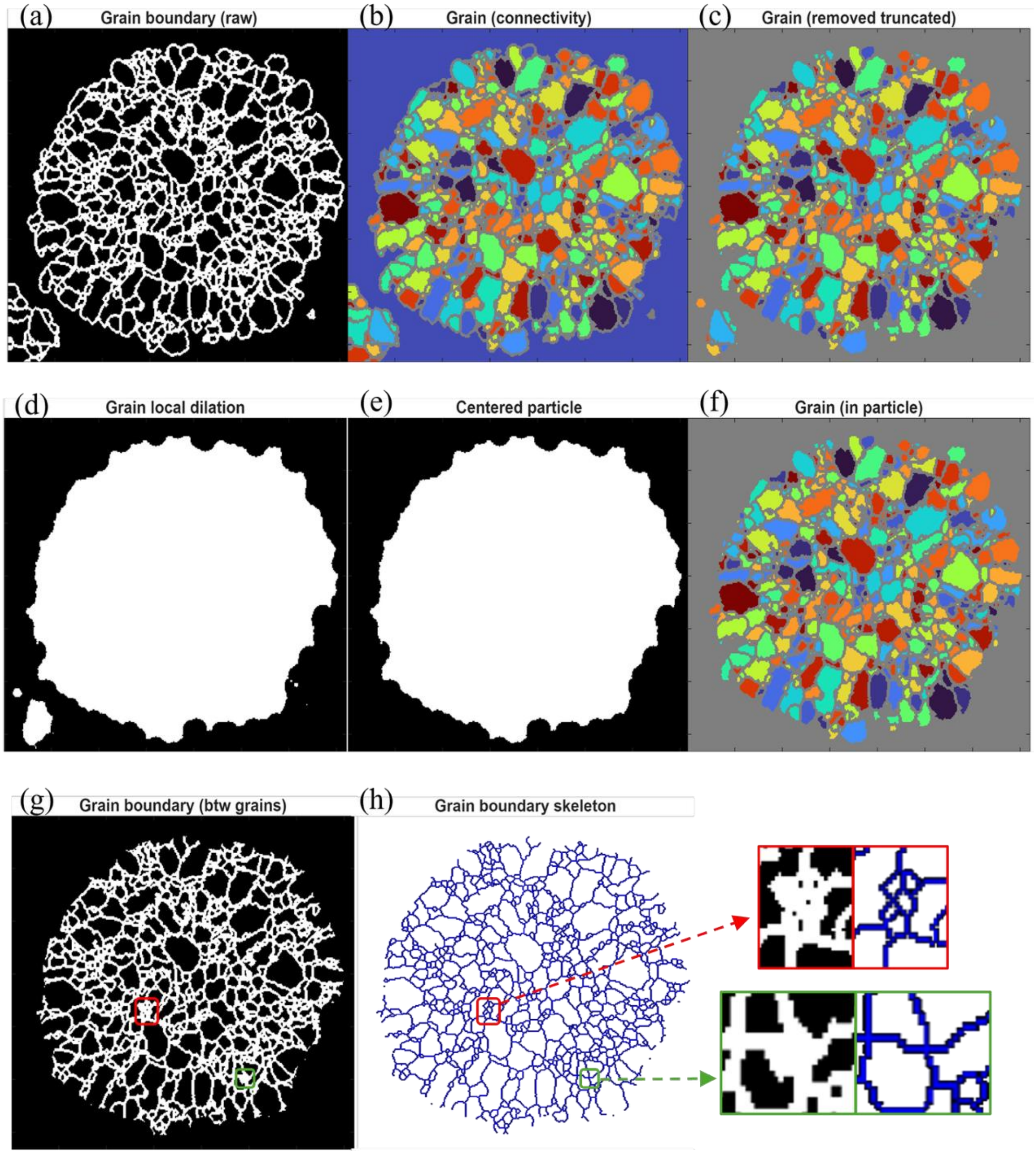

Figure 4. Grain and grain boundary identification workflow. From top left to bottom right: (a) raw grain boundary map, (b) black domain clusters (i.e., connectivity), (c) after removing the truncated clusters, (d) cluster local dilation, (e) after keeping only the main particle at the center, (f) grains within the particle of interest (b × e), (g) grain boundaries within the particle of interest (a × e), and (h) grain boundary skeleton. Skeletonization of coarse grain boundary intersections is illustrated for two examples on the bottom right.

# Image analysis

The subscripts *i, j, and k* refer, respectively, to the pixel *i*, the grain *j,* and the particle *k*. The number of pixels and grains within a particle *k* is noted, respectively, $N_{p,k}$ and $N_{g,k}$. The number of particles is noted $N_P$.

## *Image comparison (context-agnostic, before segmentation)*

This section details generic metrics typically used to compare images which are not necessarily related to the battery field. All these metrics are defined per image (that is per particle *k*) and are a comparison between the ground truth high resolution image (superscript *gt*) and the low-resolution image upscaled using one the method investigated in this work (superscript *up*). Therefore, that can only be calculated for the upscaled images using the ground truth as a reference image. The grey level value of pixel *i* in particle *k* is noted $x_{i,k}$.

- Normalized root mean squared error (NRMSE): A typical metric used in many regression-type problems that provides a pixel-per-pixel comparison. For this study, we normalize the error according to dynamic range of the image to aggregate the errors across all of the tested particles more easily (see Eq. 2).

$$NRMSE_k^{up} = \frac{1}{\max\left(x_{i,k}^{gt}\right) - \min\left(x_{i,k}^{gt}\right)} \sqrt{\frac{\sum_{i=1}^{N_{p,k}} \left(x_{i,k}^{up} - x_{i,k}^{gt}\right)^2}{N_{p,k}}} \quad [2]$$

- Peak signal-to-noise ratio (PSNR): It is defined as the logarithm of the ratio between the maximum range of the data and the reconstruction error (see Eq. 3). PSNR is usually expressed as a logarithmic quantity as signals may have a very wide range. The signal is interpreted as being the ground truth data, and the noise is the error introduced by the upscaling process. The python scikit-image built-in function *peak_signal_noise_ratio* was used with default parameters.

$$PSNR_k^{up} = 10log_{10}\left(\frac{\max\left(x_{i,k}^{gt}\right) - \min\left(x_{i,k}^{gt}\right)}{\sum_{i=1}^{N_{p,k}} \left(x_{i,k}^{up} - x_{i,k}^{gt}\right)^2 / N_{p,k}}\right) \quad [3]$$

- Structural similarity index measure (SSIM): It captures the perceived similarities between two images by comparing the luminance, contrast, and structures (correlation of pixel patterns), which are denoted by $l$, $c$, and $s$, respectively. These quantities are weighted by exponents $\alpha$, $\beta$, and $\gamma$, and use regularization constants $c_1$, $c_2$, and $c_3$, respectively. This is shown in Eq. 4 with $x$ and $y$ the upscaled and ground truth local images, $\mu_x$, $\mu_y$, $\sigma_x$, $\sigma_y$ and $\sigma_{xy}$ the local means, standard deviations, and cross-covariance).[38] SSIM ranges from -1 to 1, where -1 indicates perfect anti-correlation, 0 no similarity, and 1 perfect similarity. SSIM is calculated using a moving window and is then averaged on the whole image to provide a single value. The python scikit-image built-in function *structural_similarity* was used with default parameters.

$$SSIM_k^{up} = l(x,y)^{\alpha} c(x,y)^{\beta} s(x,y)^{\gamma} \quad \text{with} \quad \begin{cases} l(x,y) = \dfrac{2\mu_x\mu_y + C_1}{\mu_x^2 + \mu_y^2 + C_1} \\ c(x,y) = \dfrac{2\sigma_x\sigma_y + C_2}{\sigma_x^2 + \sigma_y^2 + C_2} \\ l(x,y) = \dfrac{\sigma_{xy} + C_3}{\sigma_x\sigma_y + C_3} \end{cases} \quad [4]$$

- Normalized mutual information (NMI): It quantifies the information shared between two images based on the statistical dependence of their distributions of pixel values. Exact definition is provided in.[39] It ranges from 1 (perfectly uncorrelated image values) to 2 (perfectly correlated image values, whether positively or negatively). The python scikit-image built-in function *normalized_mutual_information* was used with default parameters.

Note that lower values correspond to better performance for the NRMSE, while higher values correspond to better performance for the other metrics. SSIM and NMI metrics have finite bounds unlike NRMSE and PSNR: NRMSE> 0, PSNR $\in ]-\infty, +\infty[$, SSIM $\in [-1,1]$, and NMI $\in [1,2]$.

## *Image quality (NMC-specific, after segmentation)*

This section details quality metrics specific for segmented images with grains and grain boundaries both labelled. It is defined per image (that is per particle *k*).

In a 2D cross section representation, grain boundaries are a union of lines (excluding the unlikely case for which a grain boundary plane would be perfectly aligned with the point of view and localized exactly on the same depth), while grain and particle areas are, by definition, a surface. However, due to the digitalization of the EBSD measurement, both are represented with pixels, that is surface elements. The grain boundary width in this pixel discretization is expected to be in the same order of the pixel length (the actual width would be the true resolution of the imaging, that is ideally the pixel length but is usually larger). That is, for an infinite image resolution the cumulative area assigned to all the grain boundaries would be infinitely smaller than the area assigned to the whole particle (grain area plus grain boundary area). Therefore, a simple metric to estimate if the resolution is high enough is to calculate the ratio between these two areas (see Eq. 5, with $\Omega_{GB,k}$ and $\Omega_{G,k}$, respectively, the grain boundary and the grain domains for the particle *k*). As defined, $IQ_k$ ranges from 0 (most pixels belong to the grain boundaries: low image quality) to 1 (most pixels belong to the grains: high image quality). The image quality average among all the investigated particles, for a given image resolution and upscale method, is noted $\langle IQ \rangle$ (see Eq. 6).

|  |  |
|---|---|
| $IQ_k = 1 - \frac{\sum_{i=1}^{N_{p,k}} GB(x_i)}{\sum_{i=1}^{N_{p,k}} GB(x_i) + G(x_i)} \in [0,1] \quad \text{with} \begin{cases} GB(x_i) = 1 & \text{if } x_i \in \Omega_{GB,k} \\ GB(x_i) = 0 & \text{otherwise} \\ G(x_i) = 1 & \text{if } x_i \in \Omega_{G,k} \\ G(x_i) = 0 & \text{otherwise} \end{cases} \quad \forall k \in [1, N_P]$ | [5] |
| $\langle IQ \rangle = \frac{\sum_{k=1}^{N_P} IQ_k}{N_P}$ | [6] |

Grain boundaries width $W_{i,k}$ is calculated using the continuum particle size distribution (c-PSD), or maximum inscribed sphere method.[40] For each pixel of the grain boundary, the diameter of the largest sphere that contains it and does not overlap with the complementary domain of the grain boundary (${\Omega_{GB,k}}^C$) is assigned to it (see Eq. 7, with $S_d$ a disc of diameter *d*). Applied to the channel-like domain such as the grain boundaries, this method provides the channel width (cf. Fig. 5a). As discussed in the previous paragraph, the grain boundary width should be in the same range of the pixel length, while higher widths would indicate a degraded image. The grain with map $W_{i,k}$ is averaged to provide a mean width for each particle k, $\langle W \rangle_k$ (see Eq. 8). Lastly, the average grain width among all the particles $\langle\langle W \rangle\rangle$ is obtained using Eq. 9.

|  |  |
|---|---|
| $W_{i,k} = d \Leftrightarrow \begin{cases} x_i \in S_d \text{ and } x_i \in \Omega_{GB,k} \\ \text{and } \nexists d' > d \text{ such as } x_i \in S_{d'} \end{cases} \quad \forall i \in [1, N_{p,k}] \text{ and } \forall k \in [1, N_P]$ with $S_d \cap {\Omega_{GB,k}}^C = \emptyset \; \forall d$ | [7] |
| $\langle W \rangle_k = \frac{\sum_{i=1}^{N_{p,k}} W_{i,k}}{N_{p,k}}$ | [8] |
| $\langle\langle W \rangle\rangle = \frac{\sum_{k=1}^{N_P} \langle W \rangle_k}{N_P}$ | [9] |

Except for pulverized particles (for which cracks effectively separate particles in smaller particles), the grain boundary network should be percolating as all grains are in contact with at least another grain. The grain boundary network connectivity is calculated (using the built-in MATLAB function *bwlabel*) and the number of clusters (ideally 1) is deduced. As well, the ratio of pixels that belong to the largest grain boundary cluster is calculated (ideally 100%).

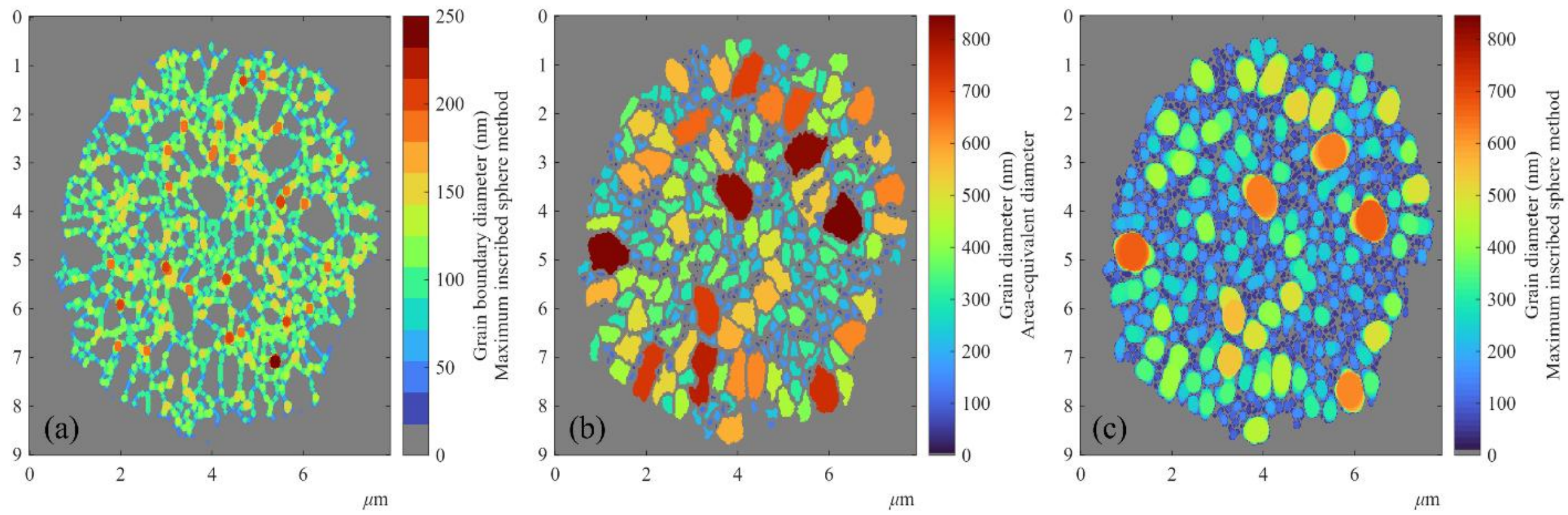

Figure 5. Illustration of (a) grain boundary width $W_{i,k}$, (b) grain area-equivalent diameter $DA_{j,k}$, and (c) grain maximum inscribed sphere diameter $DS_{i,k}$.

## *Grains and grain boundaries quantification (NMC-specific, after segmentation):*

This section details metrics specific to the particle/battery field. They provide insights into the performances and degradations of the electrode.

The area of a grain is noted $A_{j,k}$. Similarly, for any quantity $\chi$ defined per grain, its value is noted $\chi_{j,k}$. The average of $\chi$ for the particle *k* is noted $\langle\chi\rangle_k$ and deduced using a rule of mixture with $\chi_{j,k}$ weighted by the grain area $A_{j,k}$ (cf. Eq. 9). The average of $\chi$ for all the particles investigated is noted $\langle\langle\chi\rangle\rangle$, and calculated using an unweighted rule of mixture (cf. Eq. 10), and its associated 95% confidence interval, noted *CI*, is calculated (cf. Eq. 11, with r=0.025 and $t_s$ the T-score, that is CI is calculated for a +/- 2.5%). The unweighted averaging is justified as EBSD imaging is typically centered in a single particle, and thus it allows comparing metrics calculated on a single particle with the average and CI calculated for the particle library analyzed in this work. $\langle\langle\chi\rangle\rangle$ is calculated separately for each resolution and for each upscale method. The different metrics $\chi$ used to characterize the grain size and shape are detailed below.

$$\langle \chi \rangle_k = \frac{\sum_{j=1}^{N_{g,k}} \chi_{j,k} A_{j,k}}{\sum_{j=1}^{N_{g,k}} A_{j,k}} \quad \forall k \in [1, N_P]$$ [10]

$$\langle\langle \chi \rangle\rangle = \frac{\sum_{k=1}^{N_P} \langle \chi \rangle_k}{N_P}$$ [11]

$$CI = t_s(r, N_P - 1)\frac{std(\langle\chi\rangle)}{\sqrt{N_P}} \quad \text{with} \quad std(\langle\chi\rangle) = \sqrt{\frac{1}{N_P - 1}\sum_{k=1}^{N_P} (\langle\chi\rangle_k - \langle\langle\chi\rangle\rangle)^2}$$ [12]

- Area-equivalent diameter (DA*)*: For each grain *j* that belongs to a particle *k*, a diameter $DA_{j,k}$ is calculated using a disc-equivalence approach (cf. Eq. 12). This metric is illustrated in Figure 5b.

$$DA_{j,k} = 2\sqrt{\frac{A_{j,k}}{\pi}} \quad \forall j \in [1, N_{g,k}] \text{ and } \forall k \in [1, N_P]$$ [13]

- Continuous grain size distribution (DS*)*: The same expression defined in Eq. 6 is reused, except it is applied on $\Omega_{G,k}$ instead of $\Omega_{GB,k}$. This diameter represents a characteristic distance that controls the (de)lithiation dynamics within each grain. However, macroscopic Pseudo two-dimensional (P2D) battery models, based on the framework proposed by Doyle and Newman[41], use the particle diameter to model the (de)lithiation dynamics within each particle, and not the grain diameter. Nevertheless, modelers often fit the particle diameter in a P2D to match experimental data, which results in lower diameters[42] compared to what is calculated for the particle from imaging.[43] For NMC, this can be explained by electrolyte infiltration within the grain boundaries. That is, the lower bound for the NMC diameter to use in a P2D model is the primary grain average diameter.[44] This lower bound diameter is only relevant if particle inter-grain cracking is high enough to have wide enough cracks, and open-porosity cracks, so that electrolyte infiltration can happen. Unlike the area-equivalent diameter DA, DS is not calculated per grain, but per pixel. For non-circular grain DS<DA. This metric is illustrated in Figure 5c.
- Diameter distribution: In addition to calculating the average grain diameter per particle, for all the particles, the cumulative function and probability density function are also calculated for $DA_{j,k=1}$, using the grain area $A_{j,k=1}$ as weights, and for $DS_{j,k=1}$. The grain diameter distribution is only performed for a single particle for simplicity. It provides a higher granularity for the comparison between the different upscale methods.

- Grain circularity ($\xi$): The shape that has the minimum specific surface area for a given volume is the sphere. Similarly, the shape that has the minimum specific perimeter length for a given area is the disc. Therefore, the area/perimeter ratio indicates the proximity of a 2D shape with a perfect disc. The roundness $\Theta$ is used to quantify this proximity and is scaled accordingly so that it is equal to one when applied to a disc, and lower otherwise. Here, the grain circularity $\xi_{j,k}$ is calculated using the built-in MATLAB function *regionprops* that is a modification of the roundness numerically more suitable for small objects (cf. Eq. 14, with $P_{j,k}$ the grain perimeter). A high circularity (sphericity in 3D) is desirable to improve the (de)lithiation homogeneity at constant grain area (volume in 3D), as Lithium concentration within a circular (spherical in 3D) particle is invariant with the azimuth (and also the elevation in 3D), assuming the surrounding electrolyte domain is homogeneous. P2D models assume a maximum sphericity (i.e., spherical particles) which allows to calculate Lithium concentration in the active material particles only through one axis (particle radius) and not three. On the contrary, a low circularity can be beneficial to reduce kinetic resistance as it maximizes interfacial perimeter for the same grain area. Note that for non-ideal geometries (i.e., domains with a non-zero surface roughness) such as the grain shapes of real NMC particles, circularity can be very low despite exhibiting an apparent circular shape due to a high surface roughness that increases the perimeter. However, in this work, this systematic underestimation is solved as circularities are normalized with the value calculated for the initial resolution.

| $\xi_{j,k} = \Theta_{j,k}\left(1-\frac{0.5}{r}\right)^2 \in [0,1] \quad \text{with} \quad \begin{cases} \Theta = \frac{4\pi\, A_{j,k}}{{P_{j,k}}^2} \\ r = \frac{P_{j,k}}{2\pi} + 0.5 \end{cases} \quad \forall j \in [1, N_{g,k}] \text{ and } \forall k \in [1, N_P]$ | [14] |
|---|---|

- Grain solidity ($\psi$): Solidity is defined as the ratio of the grain area over the convex area of the grain and ranges from 0 to 1. The convex area is the area of the convex hull, which is the smallest convex set that contains the grain. A set is convex if any doublets of points taken within the set can be connected with a straight line that only intersects with the domain itself and not with its complementary. Here, the grain solidity $\psi_{j,k}$ is calculated using the built-in MATLAB function *regionprops*. Convex shapes such as disc have a solidity of 1. Unlike circularity, non-circular shape can reach the maximum unit value. For instance, a square is not circular but is convex (i.e., $\xi = \pi/4 < 1, \psi = 1$). Similarly with circularity, a high solidity is desirable to improve the (de)lithiation homogeneity. Some shape with a low solidity (and low circularity) can be decomposed into smaller higher solidity shapes, which for electrode microstructure can be used to identify particles.[34] The combination of circularity and solidity provides an exhaustive characterization of the grain shape.

Lastly, grain boundaries are the preferential locations for crack initiation during cycling.[1] Grain boundary density could be evaluated directly by normalizing the grain boundary area (cf. Fig. 4g) with the particle area (cf. Fig. 4e). However, this area-based approach significantly overestimates the grain boundaries for the coarse resolutions in addition to being dimensionally incorrect. Instead, we choose to normalize the grain boundaries cumulative length using the skeleton representation. However, the length-based approach is biased by the pixel discretization of the skeleton image. That is, a pixel attributed to the skeleton can represent different lengths. To remedy this, a length-calculation algorithm based on the pattern recognition of the point of contact between adjacent grain boundaries is reused from previous work[45] and is illustrated in Figure 6.

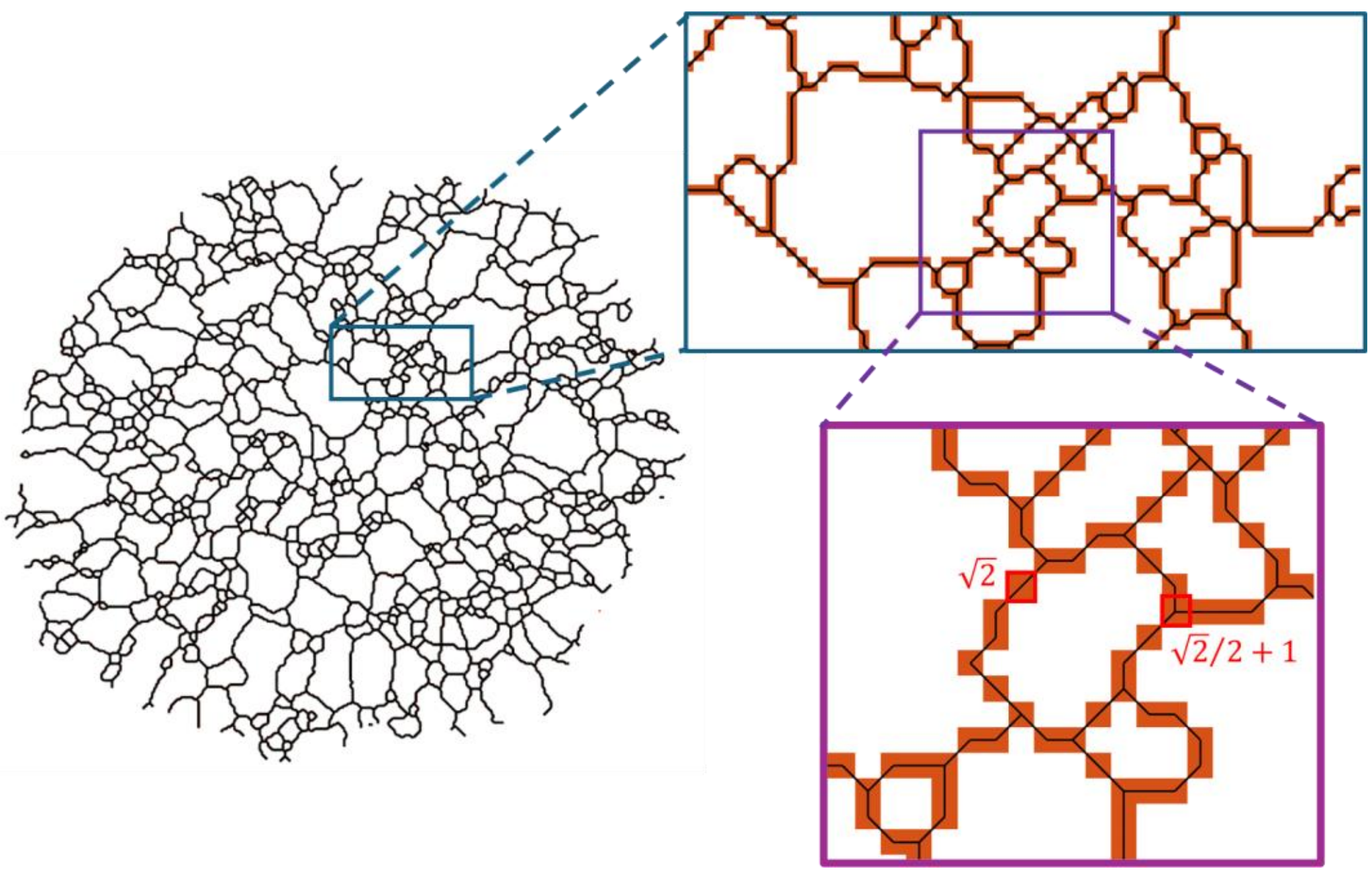

Figure 6. Grain boundary length calculation from the skeleton representation. Lengths are expressed in voxel length for this illustration.

## Results & Discussion

## Super resolved image comparison

In this section, we examine the trade-offs between the level of resolution enhancement from 2× to 12× and the quality of the resulting images. We compare the performance of the SRGAN-based enhancement against several classical approaches, including nearest neighbor, linear, and cubic interpolation as well as an unsharp masking approach, which attempts to enhance image data while preserving hard edges and textures. We aggregate metrics for the quality of the enhancement over all 15 particles for robustness. Since the SRGAN model requires training data, we train 15 distinct SRGAN models with each model holding out a different particle for testing. Each model is trained for 2,000 epochs using the Kestrel high-performance computing system at the National Laboratory of the Rockies.

Figures 7 and 8 provide qualitative comparisons between the resulting band contrast and boundary images, respectively, from the various resolution enhancement methods for a representative particle alongside the true high-resolution image. The band contrast images in Figure 7 show that every method studied here produces relatively high-quality images at a 2× resolution enhancement. For a 4× enhancement, the classical methods begin to show non-physical artifacts, such as pixelization or blurring, while the SRGAN image continues to maintain a high quality. By 8×, all grain structure has been lost in the classical approaches. The 8× SRGAN output still manages to produce reasonable grain structures, although some of the smaller grains have lost their shape or disappeared entirely. For the boundary data in Figure 8, we see that the classical super-resolution methods begin to falter as early as 2×. While these methods still produce grain structures in the super-resolved data, the boundaries between the grains are uncharacteristically thick and the grains themselves have already lost some of their actual shape. Similar to the band contrast data, all grain structure has been lost by 8× using these methods. The SRGAN approach manages to produce reasonable boundary data for 2× and 4× enhancements and the 8× enhancement still manages to produce decent reconstructions of the high-resolution boundaries. Further, comparing the band contrast data in Figure 7 and the boundary data in Figure 8, we can see that the SRGAN method produces consistent structures that match grains with boundaries between the two images. Conversely, there is little or no correlation in the structures obtained between the two images for the traditional super-resolution approaches. This highlights the added value of generating both super-resolved images simultaneously using the SRGAN approach.

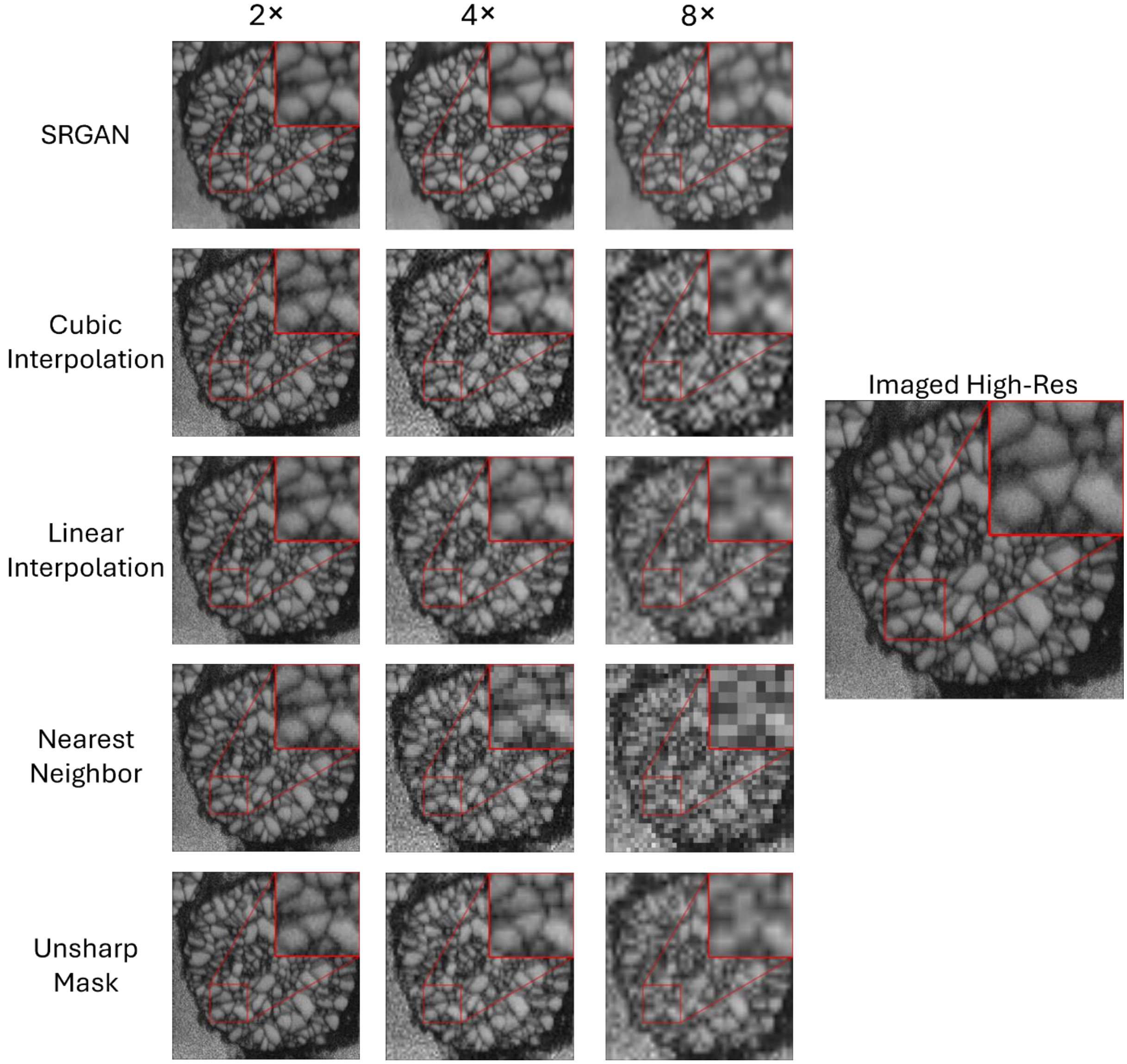

Figure 7. Side-by-side comparison of the super-resolved band contrast images using various methods for a 2×, 4×, and 8× enhancement. Red boxes provide a zoomed-in view of the particle grain structure. The imaged high-resolution particle is shown on the right for comparison.

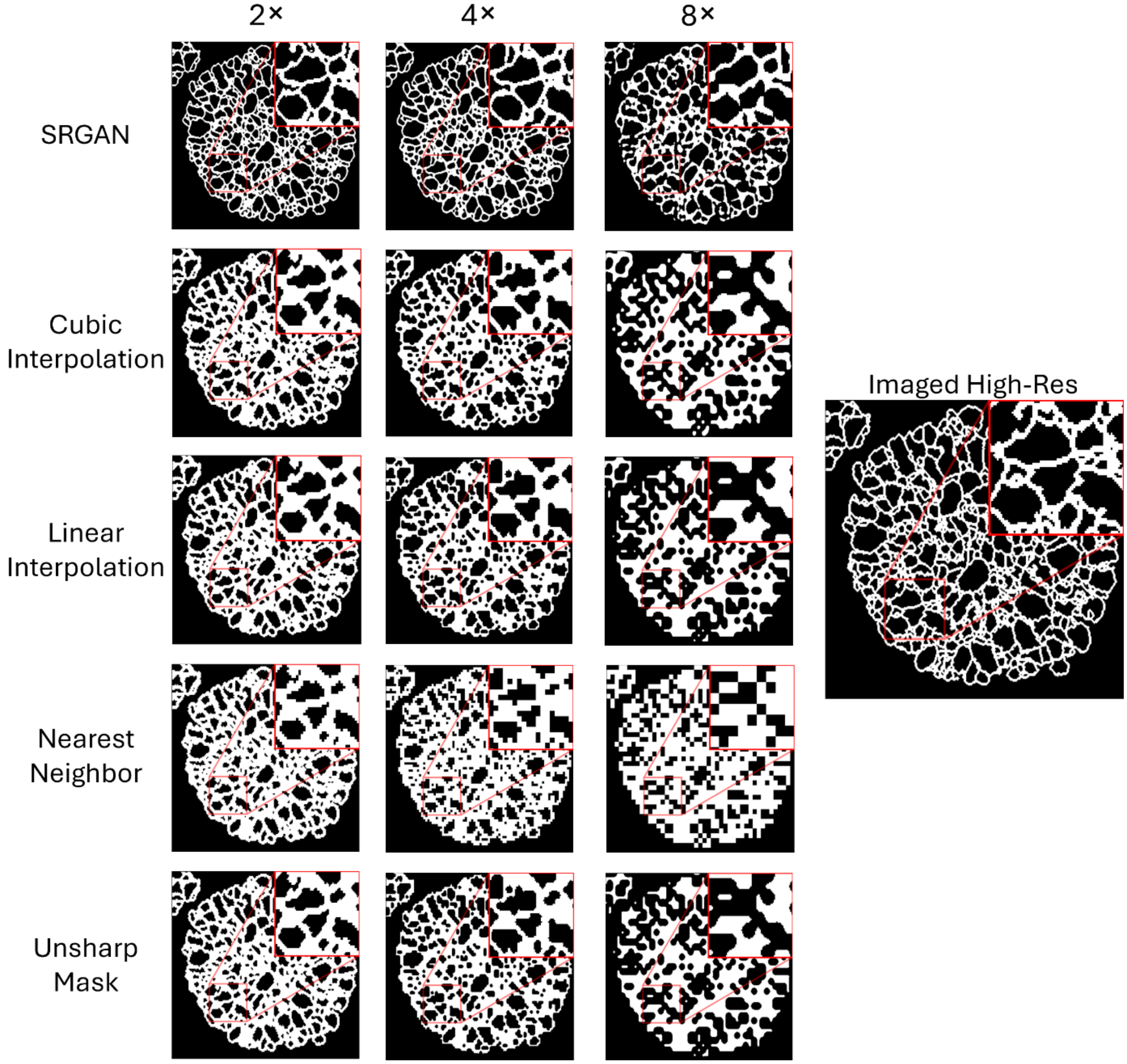

Figure 8. Side-by-side comparison of the super-resolved boundary data using various methods for a 2×, 4×, and 8× enhancement. Red boxes provide a zoomed-in view of the particle grain structure. The imaged high-resolution particle is shown on the right for comparison.

## Generic image metrics

Next, we provide a more quantitative examination of the various super-resolution methods across the range of enhancements. These analyses include both image-based metrics, which focus on the enhanced band contrast data, and morphological criteria, which examine the grain structures obtained from the boundary data. The pixel-wise metrics we consider are the normalized root mean squared error (NRMSE), the peak signal-to-noise ratio (PSNR, the structural similarity index (SSIM), and the normalized mutual information (NMI). The NRMSE is a typical metric used in many regression-type problems. For this study, we normalize the error according to dynamic range of the image in order to enable us to aggregate the errors across all of the tested particles more easily. The PSNR is a popular metric for image analysis studies. It is defined as the logarithm of the ratio between the maximum range of the data and the reconstruction error. The SSIM captures the perceived similarities between two images by comparing the luminance, contrast, and structures. NMI quantifies the information shared between two images based on the statistical dependence of their distributions of pixel values. Figure 9 shows the averages of these metrics across the various methods and resolution enhancements. Note that lower values correspond to better performance for the NRMSE, while higher values correspond to better performance for the other metrics.

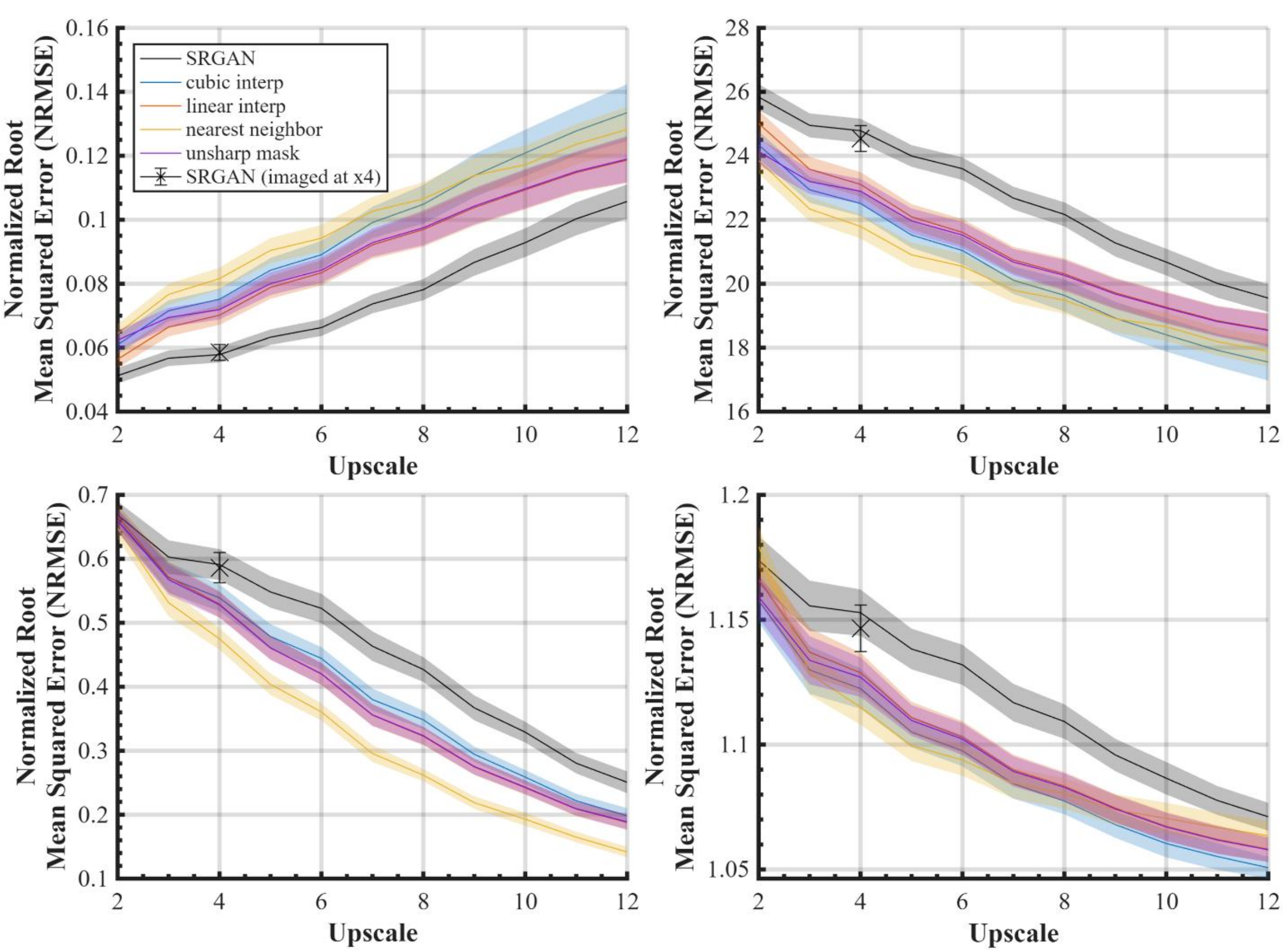

Figure 9. The average of various image-based metrics for each of the super-resolution methods as a function of the level of resolution enhancement. Note that we also provide the result of applying the SRGAN approach to the imaged low-resolution data for the 4x enhancement, denoted by the × in each plot.

Figure 9 shows mean and standard deviation of these image-level metrics for each of the super-resolution approaches across the range of enhancement levels we considered. Across all metrics, we see that the SRGAN outperforms the classical methods for nearly every level of enhancement. At a 2× resolution enhancement, the performance converges slightly across the methods, especially for the SSIM. For the small amount of resolution enhancement being performed here, interpolation is a reasonable approach. However, as the resolution enhancement increases, the SRGAN quickly stands out as the best approach. Notably, there is no specific level of enhancement at which the super-resolution process suddenly drops off despite the clear breakdown in grain structures that we observed in the qualitative results from Figure 7. This highlights the limitation of the image-level metrics to characterize the quality of physically relevant features in data. We also include the result from applying the real imaged low-resolution data to the trained SRGAN model to further validate the downscaling process discussed earlier. Here, we see that the enhanced output conditioned on the imaged low-resolution data aligns well with the quality of the results obtained using the artificially downscaled images at the same enhancement level. This consistency provides confidence that the proposed methodology is robust and can be deployed in real workflows.

## Image quality

The image quality metric (cf. Eq. 5,6) defined specifically for segmented particles with their grain boundaries identified is shown in Figure 10a. The ground truth images have on average an image quality of 0.645 +/- 0.019 (95% confidence interface). SRGAN systematically significantly outperforms all the other upscaling methods considered in this work. With SRGAN, upscaling first causes a slight decrease of the image quality. A lower plateau, ~0.59, is reached for the 5× to 7× upscale levels. This initial decrease is expected as the grain width (and thus the grain cumulative area) is overestimated at low resolution, and the upscaling can only partly retrieve the ground truth thinner with (cf. Fig. 10c). For higher upscaling levels (8× and above), the image quality is increasing and eventually surpasses the ground truth value. This increase is, however, misleading: it does not correspond to an actual improvement of the image but is due to the grain boundary domains being removed as the image is severely degraded (cf. Fig. 10d), to the point metrics defined for an NMC particle are getting meaningless. The resolution for which this degeneration of the image occurs is also indicated by the number of grain boundary clusters, and especially the relative size of the largest one. Ground truth images have roughly 10 grain boundary clusters on average, however the largest one represents more than 99% of the total grain boundary area, that is, the grain boundary network is fully connected, even when observed in a cross section (cf. Fig. 10b). SRGAN shows the share of the largest grain boundary cluster starts to decline for upscaling higher than 6×. This analysis indicates that an upscaling from a resolution up to between 125 and 150nm (that is 5× and 6×) preserves the image quality with SRGAN.

The other upscaling methods can be categorized into two groups based on these image quality metrics. First, the linear and cubic interpolations, and the unsharp mask methods provide very similar image results. Second, the nearest neighbor method provides the worst outcome. They all have in common a sharp decline in quality from the first level of upscaling (2×): from 0.645 (ground truth) down to 0.443-0.427 for the first group, and down to 0.379 for the nearest neighbor method, way lower than what is achieved with SRGAN (down to 0.624). Poor performance for the baseline methods is due to a significant coarsening of the grain boundaries (cf. Fig. 10d), which result in very wide, unrealistic, grain boundary widths (cf. Fig. 10c), and thus a low image quality.

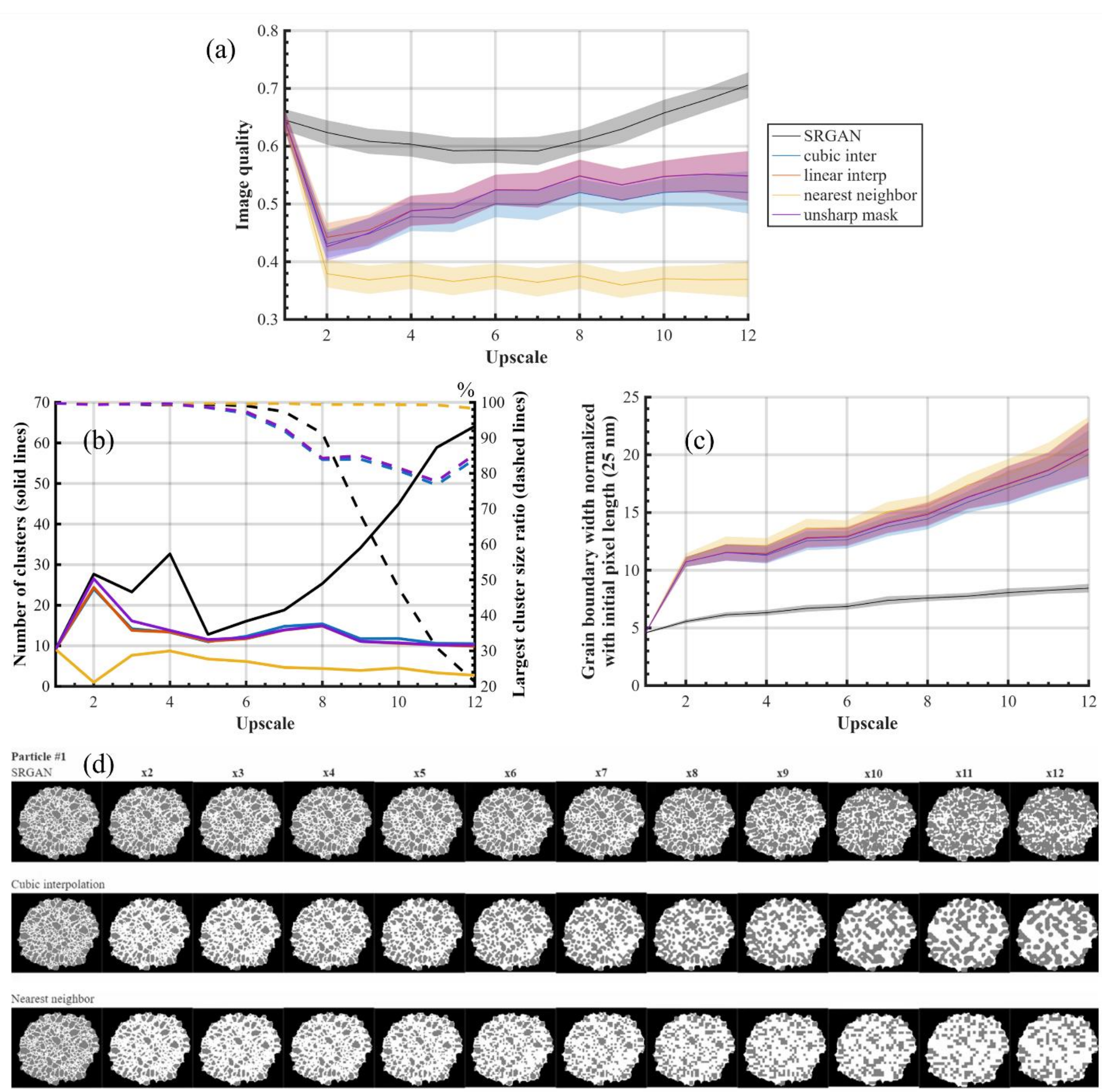

Figure 10. (a) Image quality averaged among all the particles calculated for the ground truth images and the images upscaled with different methods, for degraded resolution ranging from 2× to 12×. (b) Number of grain boundary clusters, and ratio of the largest one. (c) Grain boundary diameter (maximum sphere-inscribed method). (d) Visualization of the grain boundary coarsening and disconnection occurring at high upscaling levels (videos available in Supplementary Information show all particles, with all resolutions, and all methods).

## Quantification of morphological properties

### *Grain size and numbers:*

Figure 11a shows the number of grains normalized with the ground truth images for the different upscaling methods. All methods, except for SRGAN, exhibit a sharp reduction with the first downscaling-upscaling level (2×), retaining ~48% of the initial number of grains. On the contrary, SRGAN managed to preserve 80% of the grain number at this resolution. SRGAN's advantage is systematic for all the resolutions investigated, with for instance a 44% retention at 6× compared to 16% for all the other methods. Among the baseline methods considered in this work, the nearest neighbor approach is the worst, while the three other approaches (linear and cubic interpolations, and unsharp mask) are roughly similar.

Despite SRGAN losing 20% of grains at the first upscaling level, the penalty is much more contained for the grain average diameters: only 5.1% and 1.02% higher, respectively, for the maximum sphere-inscribed method (cf. Fig. 11b) and the area-equivalent method (cf. Fig. 11c,d). This indicates that most of the grain information lost during the downscaling-upscaling scheme is concentrated on the smallest grains. These small features correspond either to the cross sections of small grains or out-of-plane tips of larger grains. If a 10% degradation threshold is considered, SRGAN preserves the maximum sphere-inscribed diameter up to a 6× scaling (+8%) and the area-equivalent diameter up to a 5× scaling (+5.5%). Above these scaling, the error is above 10% and degrades further as the grain boundary disconnection occurring at these degraded resolution result in larger grains (cf. Fig. 10b,d).

While SRGAN tend to systematically overestimate the diameters, the other methods first underestimate them before eventually also overestimating them. This different trend is due to the grain boundary coarsening, absent in SRGAN, but systematic in the other approaches (cf. Fig. 10c,d). Indeed, as the boundaries are getting wider, the grains in between are shrinking, and the shrinking of the large grain overshadow the removal of the smallest grains. Additionally, the higher the upscaling is, the higher are the variations between each upscaled particles for all the methods considered in this work.

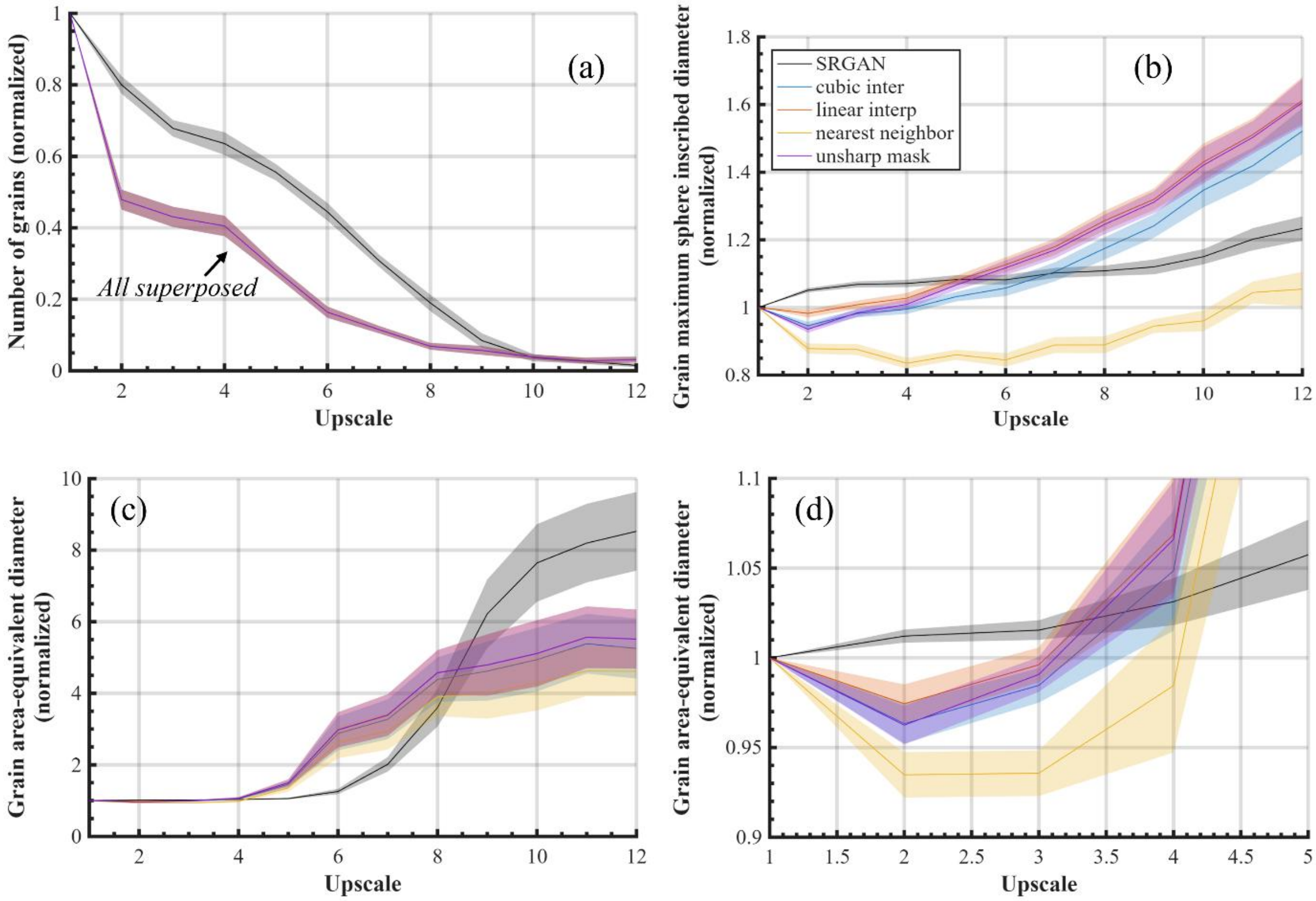

Figure 11. (a) Number of grains, grain average diameter calculated with (b) the maximum inscribed sphere method and (c, d) the area-equivalent method, for images upscaled with different methods, for degraded resolution ranging from 2× to 12× and normalized with the values of the ground truth image.

To further evaluate the loss of information induced by the coarse resolutions, the grain size distributions are plotted for a single representative dataset for all the methods in Figure 12 for a selection of resolutions. Up to 4×, all methods show similar area-equivalent diameters (cf. Fig. 12, top row), except for the small diameters for which SRGAN outperforms the other upscaling approaches. At 5×, SRGAN is still relatively close with the ground truth while all other methods start to deviate significantly in favor of larger diameters. At 6×, SRGAN is significantly off. At 7× and above, the grain boundary network is losing its connectivity (cf. Fig. 10b) and the grains are merging even more (cf. Fig. 10a), which results in unrealistic large grain area-equivalent diameters. The maximum sphere inscribed diameter distributions are much better preserved compared to the area-equivalent diameter distributions, even at very high upscaling (cf. Fig. 12, bottom row). This is because the sphere inscribed method is hardly affected by the grain boundary percolation. The main issue lies in the strong underestimation of the grain volume assigned with the small diameters, systematic for all methods but especially substantial for the linear and cubic interpolation approaches. Here again, SRGAN provides the best estimation of the small diameters among all the methods investigated. Figure 13 shows the cumulative and probability density functions for the same data set, but only for SRGAN. The area-equivalent diameters distributions are preserved up to a 5× upscaling. The maximum sphere inscribed diameter distributions are mostly a rebalancing between the smaller and the larger diameters.

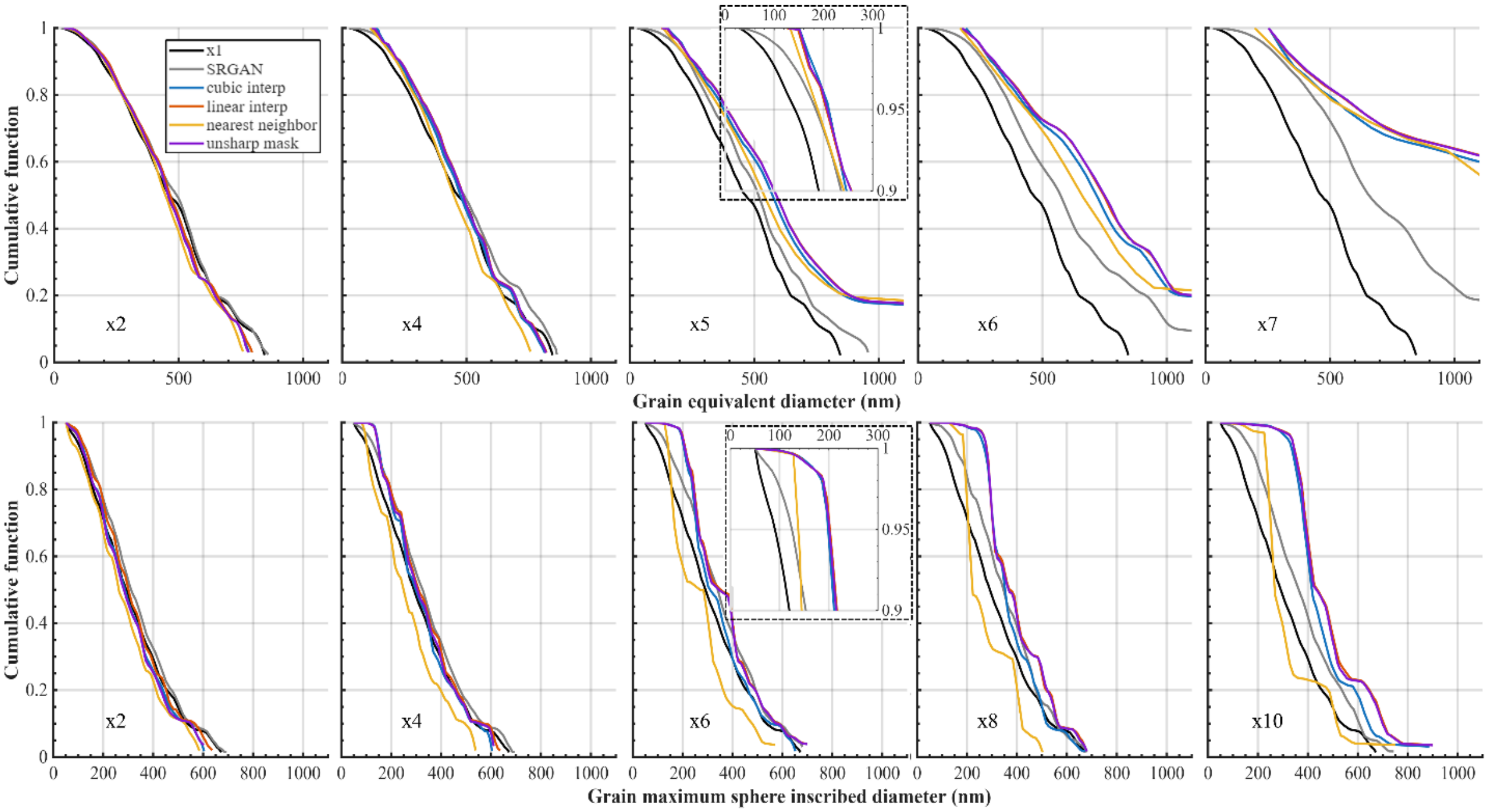

Figure 12. Cumulative functions of (top row) grain area-equivalent diameter and (bottom row) grain maximum sphere inscribed diameter calculated for a representative dataset, for the ground truth image and for images upscaled with all the upscaling methods. Inserts show the distributions for the 10% grain domain associated with the smallest diameters.

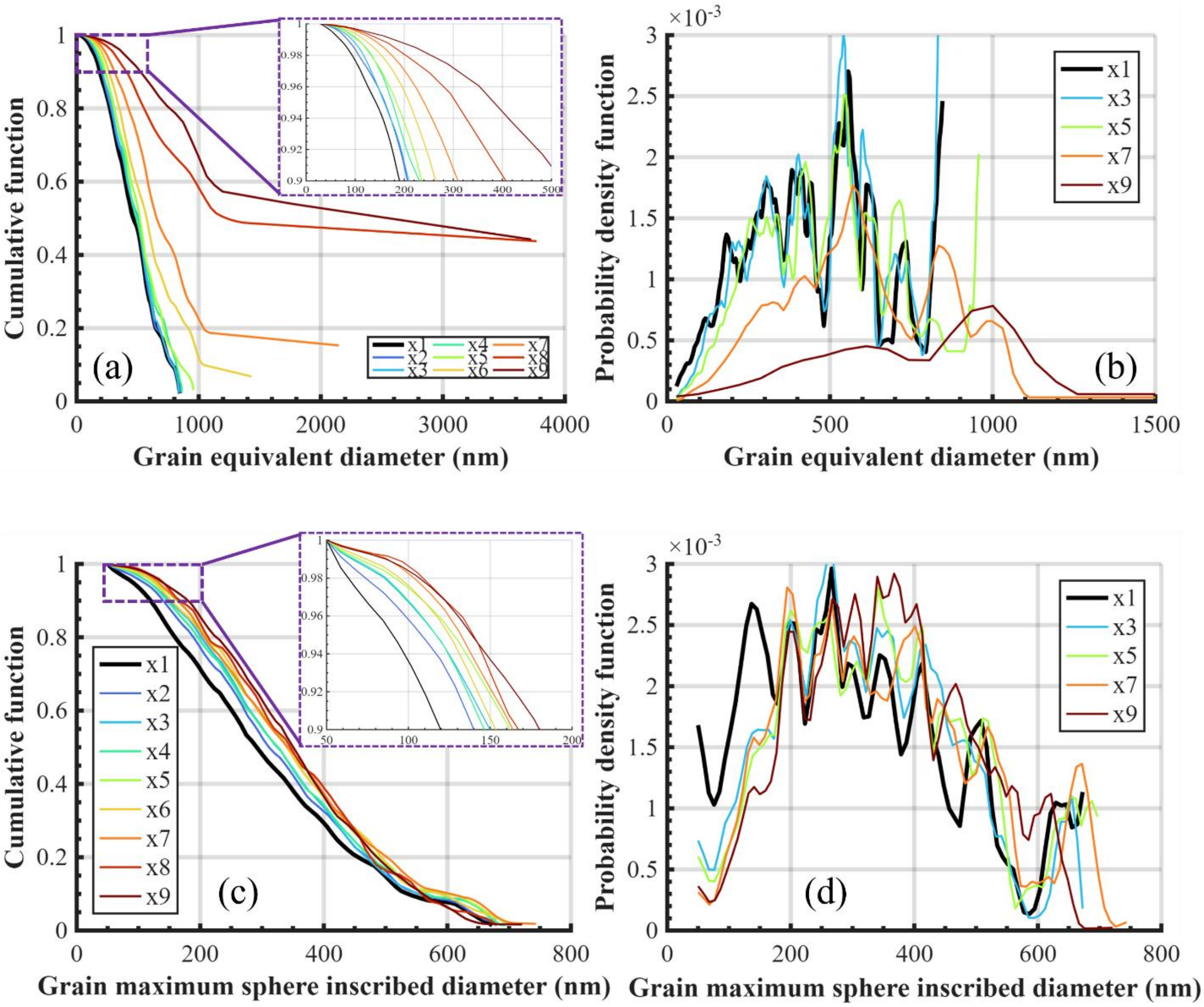

Figure 13. Grain area-equivalent diameter (a) cumulative functions and (b) probability density functions, and grain maximum sphere inscribed diameter (c) cumulative functions and (d) probability density functions, calculated for a representative dataset, for the ground truth image and for images upscaled with the SRGAN method only. (a, c) Inserts show the distributions for the 10% grain domain associated with the smallest diameters.

*Grain shape:*

Figure 14 shows the grain shape metrics calculated for both methods and normalized with the ground truth images. SRGAN preserves shape (both circularity and solidity with an error below ~10%) for an upscaling up to 5× to 6×. SRGAN slightly overestimates these two metrics (i.e., the methods tend to recreate slightly more convex grain shapes). Above a 6× scaling, solidity starts to degrade significantly and circularity variability increases. The sharp decrease calculated for solidity is coherent with the associated sharp increase in the grain area equivalent diameter (cf. Fig. 11c) and the grain boundary connectivity loss (cf.

Fig. 10b). Indeed, as the grain boundary network is losing its percolation, adjacent grains previously separated by a grain boundary get in contact which results in a new grain, much larger (area-wise) and with a highly concave shape. For low to intermediate upscaling (up to 4×), linear and cubic interpolations and unsharp methods provide a better representation of the grain shape compared to SRGAN. The nearest neighbor method provides the worst shape degradation.

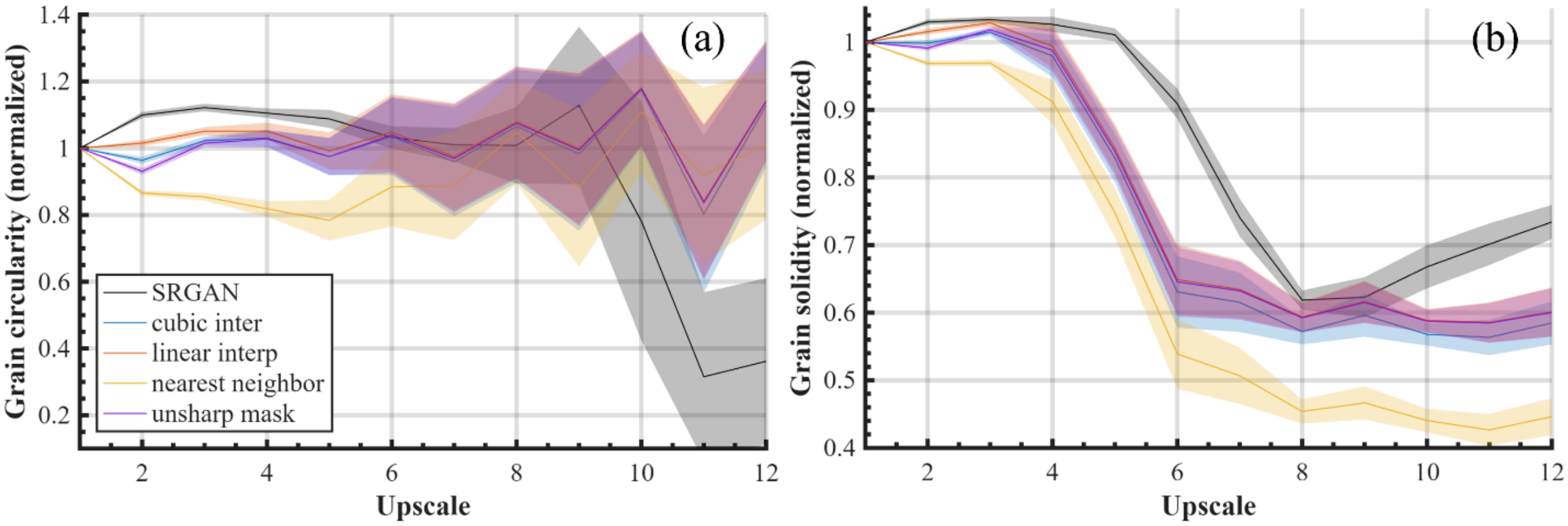

Figure 14. Grain shape analysis performed with two metrics, (a) grain circularity and (b) grain solidity, calculated for images upscaled with different methods, for degraded resolutions ranging from 2× to 12×, and normalized with the ground truth image. Note that both metrics can go higher than 1 as the values reported here are normalized with the values calculated for the ground truth images.

### *Grain boundary length density:*

Figure 15 shows the grain boundary cumulative length density calculated for all methods and normalized with the ground truth images. A significant reduction is calculated for all approaches as the number of grains is decreasing with the upscaling level (cf. Fig. 11a). The reduction is also expected as surface roughness increases with finer resolutions. SRGAN outperforms the other methods (except for very high upscaling, above 8×). SRGAN manages to get 93.0% and 80.1% of the initial grain boundary length density, respectively, at 2× and 6×. If a 10% degradation threshold is considered, SRGAN preserves the grain boundary length density up to 3×. Above this scaling, the error is above 10% and degrades further linearly, with the degradation slope increasing significantly at 5× to 6×.

Among the baseline methods considered in this work, the nearest neighbor approach is the best, while the three other approaches (linear and cubic interpolations, and unsharp mask) are roughly similar. Unlike the other metrics previously investigated, the nearest neighbor method is a worthy choice, especially considering its simplicity. This is due to its excellent preservation of the grain boundary network, that stays connected even at very high upscaling (cf. Fig. 10b). Especially, the nearest neighbor method overcomes SRGAN above 8× (albeit, for such high upscaling, the error is very large, with a -40% underestimation of the ground truth value), which corresponds to the upscaling level when SRGAN no more preserves the connectivity of the grain boundary network (cf. Fig. 10b).

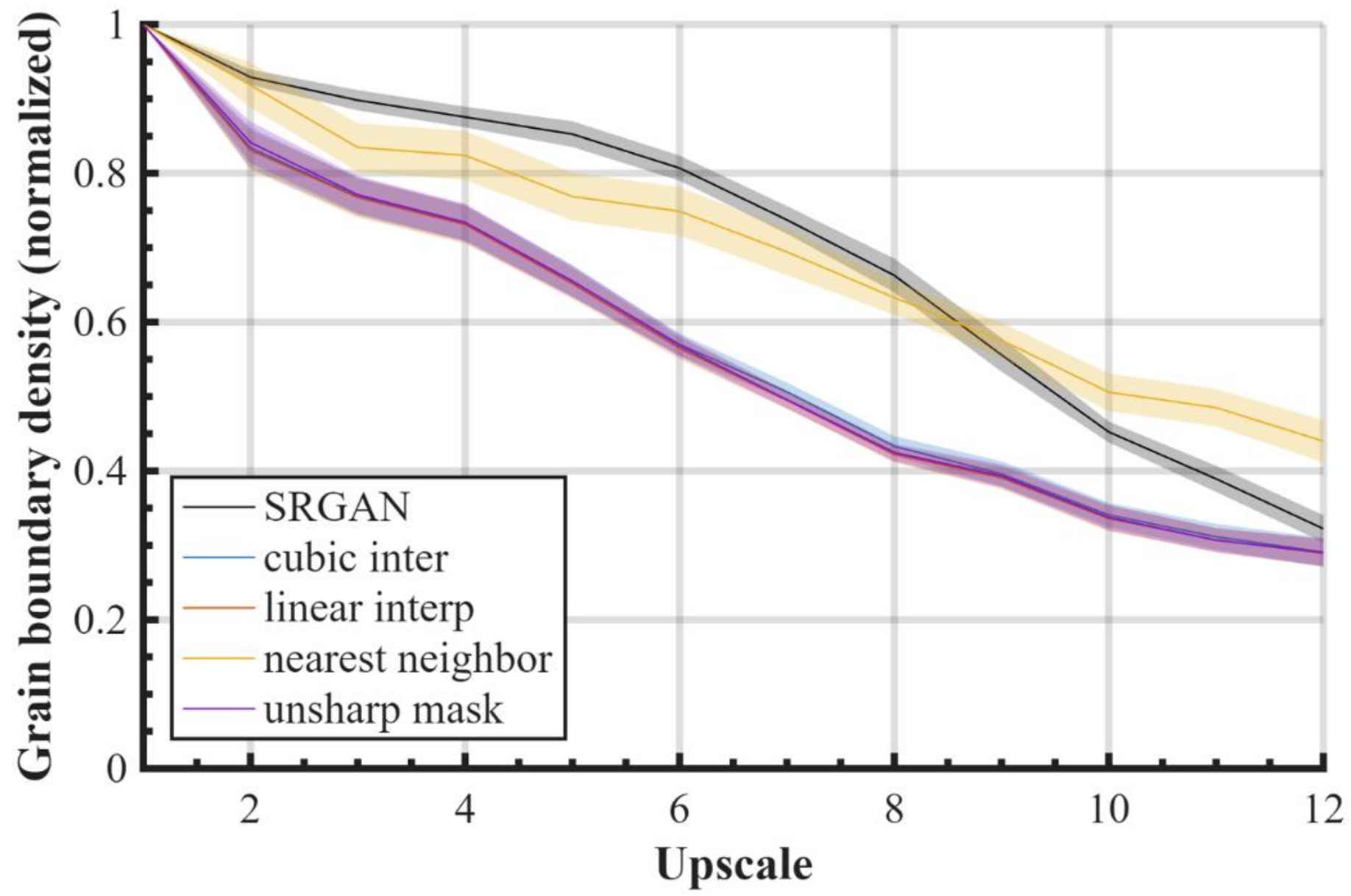

Figure 15. Grain boundary length calculated for particles upscaled with different methods, for degraded resolution ranging from 2× to 12×, and normalized with the ground truth images.

## Demonstration of high-throughput wide-area EBSD

To demonstrate the capability and throughput enhancement of the SRGAN model developed in this work, a large-area EBSD dataset was acquired at 100 nm/pixel resolution, encompassing approximately 35 cathode particles contained within an electrode cross-section. The total acquisition time for the dataset was 40 minutes. For reference, the same dataset acquired with a resolution of 25 nm/pixel would take approximately 11 hours. The performance of the SRGAN model, shown qualitatively in Figure 16, demonstrates notable enhancement in the clarity of the band contrast map along with a significant improvement in the resolution of the segmented grain boundary map, which is critical for extracting quantitative microstructural metrics from these EBSD datasets. The time savings afforded by the SRGAN framework enables the acquisition and analysis of a substantially greater number of data points, and therefore microstructural features, thereby providing more robust and statistically accurate representations of the material being studied in less time than traditional methods.

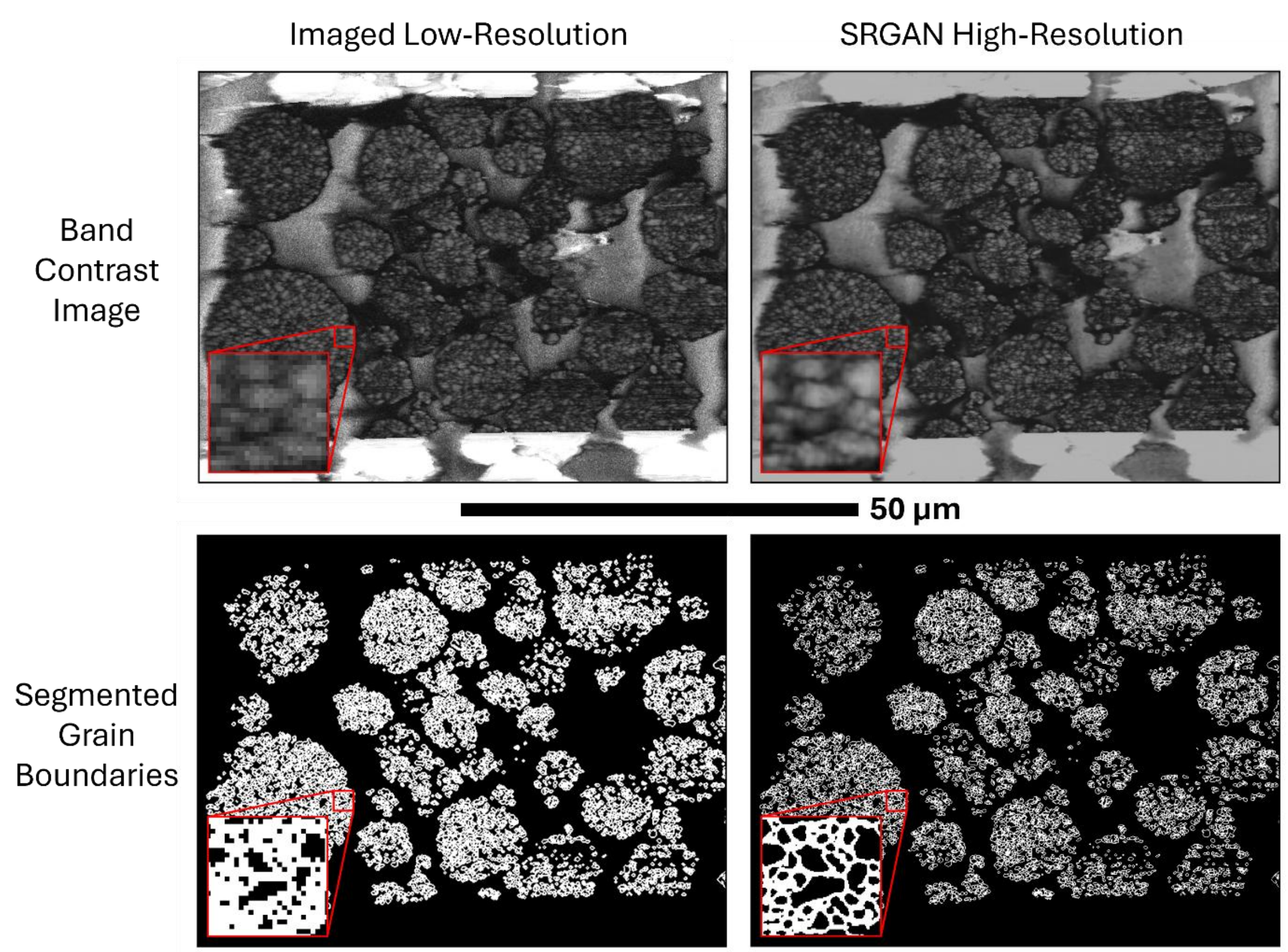

Figure 16. Representative large field of view cross-sectional EBSD map of the same NMC material demonstrating the SRGAN methodology developed in this work. Qualitative comparison between the band contrast (top row) and grain boundary (bottom row) maps of imaged low-resolution (left) and super-resolved (right) datasets demonstrates the improvement in microstructural clarity, notably in the segmented grain boundary map.

# Conclusions

This study demonstrates the application of a super-resolution generative adversarial network model trained on electron backscatter diffraction imaging data to significantly improve throughput of EBSD characterization. The developed SRGAN model was applied to 15 EBSD datasets across super-resolution upscaling factors ranging from 2× to 12× and compared to classical super-resolution methods, such as linear interpolation. Quantitative microstructural analysis verified that the SRGAN systematically outperformed the other methods considered in this work. Especially,

SRGAN better preserves the small grains, the thin width of the grain boundaries, and the grain boundary network percolation. For an arbitrary 10% error threshold, SRGAN preserves grain size and shape up to a 5x upscaling, and the grain boundary length density up to a 3x upscaling. The latter being more penalized as surface roughness is poorly described at low resolutions. Since the grain size is related to the characteristic diffusion length used in P2D representations to model the (de)lithiation dynamics within the active material particles (assuming electrolyte infiltration within the grain boundaries and/or cracks), it can be considered that the errors are acceptably constrained up to 5× upscaling. That is, the model systematically predicts reasonably accurate metrics across all datasets for SRGAN images generated from 125 nm/pixel resolution images when compared to corresponding 25 nm/pixel ground truth images. These results highlight the potential of deep learning–based GAN models to significantly reduce acquisition times (25× faster) and/or increase practical fields of view (25× larger) without substantial loss of quantitative accuracy.

Beyond the demonstrated case, this methodology is broadly adaptable to other material systems and EBSD modalities where high temporal or spatial throughput is critical. Potential applications include accelerated mm-scale characterization of hydrogen fuel cell electrode architectures, thin-film photovoltaic microstructures (e.g., CdTe), and complex microstructures of additively manufactured components. In situ EBSD experiments stand to benefit in particular, as the enhanced temporal resolution afforded by super-resolution would enable more detailed monitoring of microstructural evolution during environmental stressing. This framework would also greatly speed up the EBSD acquisition steps in 3D-FIB-EBSD workflows, which are typically the rate-limiting steps, thereby enabling larger volumes to be studied more efficiently.

In summary, this work demonstrates that the adoption of GAN-based super-resolution methods could substantially advance EBSD as a high-throughput, high-fidelity microstructural characterization tool for materials science research and industrial process monitoring.

## Acknowledgements

This work was authored by the National Laboratory of the Rockies (NLR), for the U.S. Department of Energy (DOE) under Contract No. DE-AC36–08GO28308. A portion of this research was performed using computational resources sponsored by the U.S. Department of Energy's Office of Critical Minerals and Energy Innovation and located at the National Laboratory of the Rockies. The views expressed in the article do not necessarily represent the views of the DOE or the U.S. Government. The publisher, by accepting the article for publication, acknowledges that the U.S. Government retains a nonexclusive, paid-up, irrevocable, worldwide license to publish or reproduce the published form of this work, or allow others to do so, for U.S. Government purposes.